\documentclass[final,3p,times,twocolumn]{elsarticle}

\usepackage{amssymb}
\usepackage{amsmath}
\usepackage{subcaption}
\usepackage{graphicx}
\usepackage{multirow}
\usepackage{multicol}
\usepackage{url}

\usepackage{breakurl}
\usepackage[breaklinks, hidelinks]{hyperref}
\usepackage{comment}
\usepackage{xcolor}

\begin{document}

\begin{frontmatter}

%% Title, authors and addresses
\title{GA-SmaAt-GNet: Generative Adversarial Small Attention GNet for Extreme Precipitation Nowcasting}

\author{Eloy Reulen}

\author{Siamak Mehrkanoon\corref{cor1}}
\ead{s.mehrkanoon@uu.nl}
\cortext[cor1]{Corresponding author}

\address{Department of Information and Computing Sciences, Utrecht University, Utrecht, The Netherlands}

\begin{abstract}
In recent years, data-driven modeling approaches have gained significant attention across various meteorological applications, particularly in weather forecasting. However, these methods often face challenges in handling extreme weather conditions. In response, we present the GA-SmaAt-GNet model, a novel generative adversarial framework for extreme precipitation nowcasting. This model features a unique SmaAt-GNet generator, an extension of the successful SmaAt-UNet architecture, capable of integrating precipitation masks (binarized precipitation maps) to enhance predictive accuracy. Additionally, GA-SmaAt-GNet incorporates an attention-augmented discriminator inspired by the Pix2Pix architecture. This innovative framework paves the way for generative precipitation nowcasting using multiple data sources. We evaluate the performance of SmaAt-GNet and GA-SmaAt-GNet using real-life precipitation data from the Netherlands, revealing notable improvements in overall performance and for extreme precipitation events compared to other models. Specifically, our proposed architecture demonstrates its main performance gain in summer and autumn, when precipitation intensity is typically at its peak. Furthermore, we conduct uncertainty analysis on the GA-SmaAt-GNet model and the precipitation dataset, providing insights into its predictive capabilities. Finally, we employ Grad-CAM to offer visual explanations of our model's predictions, generating activation heatmaps that highlight areas of input activation throughout the network.
\end{abstract}


\end{frontmatter}

%% main text
\section{Introduction}
\label{sec:Intro}
Precipitation nowcasting \cite{nowcasting}, also referred to as short-term precipitation forecasting, has long been a focal point of meteorological research. Especially, the ability to accurately predict short-term extreme precipitation events holds particular significance, as these weather phenomena have a substantial impact on both human life and economy. Notably, the Netherlands has experienced a noteworthy increase in precipitation over the past century, as highlighted in the findings of studies such as \cite{precip_trends_nl, kmni_extreme}. Furthermore, the frequency of extreme precipitation events in the Netherlands has been on the rise, as demonstrated by \cite{extreme_precip_nl, kmni_extreme}. In light of these developments, precise nowcasts for extreme precipitation events are of great importance for timely warnings and precautionary measures.
Traditionally, meteorologists have relied upon Numerical Weather Prediction (NWP) models for generating weather forecasts \cite{nwp}. NWP models employ intricate mathematical equations to simulate various atmospheric properties. This involves solving a set of complex partial differential equations governing the physical state of the atmosphere. While NWP models can deliver accurate forecasts, their computational demands and limitations in resolving small-scale complex atmospheric processes \cite{deep_shared} make them less suitable for nowcasting tasks which require short-term forecasts, typically ranging from minutes to a maximum of two hours.
In recent years, Convolutional Neural Networks (CNNs), a class of deep learning models particularly well-suited for image and spatial data analysis, have emerged as a promising area of research for tackling the precipitation nowcasting problem. CNNs have demonstrated remarkable success in various fields, including computer vision \cite{alex, resnet, vgg}, natural language processing \cite{kim2014convolutional, zhang2016characterlevel}, cyber security \cite{dos, fuzzycyberattack}, and, increasingly, meteorology \cite{deep_shared, convLSTM, trajGRU,  stunner, scent, trebing2020wind}. Their ability to extract spatial patterns and features from radar or satellite imagery has made them valuable tools in the quest for more accurate precipitation forecasts, even for extreme weather events \cite{dgmr}.
However, it is essential to acknowledge that CNNs, while showing promise, often encounter significant challenges when predicting extreme precipitation events \cite{extreme_hard}. These events exhibit complex and non-linear behaviors that can strain the capabilities of traditional deep learning models. As such, ongoing research in this field is not only focused on leveraging CNNs but also on enhancing their performance in predicting extreme weather events through innovative architectures \cite{dgmr}, data augmentation techniques \cite{gan_mask}, and the integration of additional meteorological variables \cite{nowcastnet}. By addressing these challenges, meteorologists aim to harness the full potential of deep learning techniques in advancing the accuracy and reliability for extreme precipitation nowcasting.

It is important to note that the precise definition of extreme precipitation can vary between different regions and research studies. For the purposes of this paper, when we refer to extreme precipitation in the Netherlands, we are specifically referring to precipitation depths exceeding 20 mm accumulated within a one-hour time frame, as most drainage systems in the Netherlands are not able to process more than this amount of rain per hour \cite{knmi_zwaar}. 

In this study, we demonstrate the effectiveness of data augmentation and a generative approach for extreme precipitation nowcasting.
To this end, we present GA-SmaAt-GNet, a novel generative adversarial framework designed to enhance the performance of the widely adopted SmaAt-UNet architecture \cite{SmaAtUnet} in the context of extreme precipitation nowcasting. Here, we briefly provide the key contributions of our work as follows: 
\begin{enumerate}
    \item We develop SmaAt-GNet which extends SmaAt-UNet by incorporating precipitation masks (binarized precipitation maps) as an additional data source. This provides supplementary information to the network and allows the model to better understand where areas of specific precipitation intensity are located. This architecture also paves the way for the integration of more data sources to enhance the predictive performance of deep learning models for meteorological tasks.
    \item We develop GA-SmaAt-GNet, a generative adversarial framework, which utilizes our proposed SmaAt-GNet as a generator. Additionally, it incorporates a novel discriminator architecture that utilizes the CBAM attention mechanism \cite{cbam}. This approach allows the discriminator to focus on critical areas in the data, improving its ability to distinguish between real and generated precipitation patterns.
    \item We delve into both epistemic uncertainty of the models and aleatoric uncertainty inherent in the precipitation data. Our analysis reveals that both forms of uncertainty escalate with higher precipitation intensities, underscoring the inherent complexities involved in the task of extreme precipitation nowcasting.
    \item We utilize Grad-CAM, a visual explanation technique, to generate heatmaps of the activations of different parts of our proposed network to gain further insight into its predictions. With these generated heatmaps we show that the precipitation maps and precipitation masks that serve as input to our proposed SmaAt-GNet architecture contribute to the distinct areas of the model's prediction.
    \item We evaluate the performance of our proposed models on a real-world precipitation dataset from the Netherlands. Spanning 25 years, from 1998 to 2022, this dataset provides a comprehensive and realistic environment to assess the model's capabilities in predicting extreme weather events. 
    \item We conduct a comparative analysis of our proposed models against other established architectures, such as SmaAt-UNet \cite{SmaAtUnet} and RainNet \cite{rainnet}. Our results reveal that both proposed SmaAt-GNet and GA-SmaAt-GNet exhibit enhanced overall performance when evaluated on the precipitation dataset as well as a notable improvement in nowcasting extreme precipitation events.
\end{enumerate}

This paper is organized as follows: we first give a brief overview of related literature on weather forecasting in Section \ref{sec:related_work}. We then introduce our proposed models in Section \ref{sec:methods}, while Section \ref{sec:experiments} describes the used dataset and conducted experiments. The results of the experiments are discussed in Section \ref{sec:results} and finally we make some concluding remarks in Section \ref{sec:conclusion}.

\section{Related work}\label{sec:related_work}

In recent years, deep learning techniques have found applications in various weather forecasting tasks \cite{deep_shared}. For instance, Shi et al. introduced ConvLSTM, which expanded upon the LSTM architecture by incorporating convolutional structures in the input-to-state and state-to-state transitions \cite{convLSTM}. Their approach demonstrated improved performance when compared to traditional optical flow-based methods for precipitation nowcasting. Building upon the success of ConvLSTM, a subsequent study introduced TrajGRU \cite{trajGRU}. TrajGRU extended ConvLSTM by employing a GRU (Gated Recurrent Unit) instead of an LSTM, a modification that enables the model to learn location-specific patterns more effectively.

More recently, U-Net-based architectures have gained traction in the field of weather forecasting. Originally designed for biomedical image segmentation \cite{unet}, the U-Net architecture has been repurposed effectively for various applications, including weather forecasting. This architecture consists of an encoder network, which progressively reduces the spatial resolution of the input image, and a decoder network that upsamples the feature maps back to the original image dimensions. These two networks are interconnected via skip connections, enabling the decoder to access high-resolution features from the encoder. While initially intended for biomedical image segmentation, the U-Net architecture has found successful application in various domains, including weather forecasting \cite{aatransunet, SmaAtUnet, sar}. Trebing et al. \cite{SmaAtUnet} extended the U-Net architecture by incorporating attention modules and depthwise-separable convolutions, successfully deploying it for precipitation nowcasting in the Netherlands and cloud cover nowcasting in France. The prediction performance of their SmaAt-UNet architecture is on par with the traditional U-Net, but with a notable reduction in the number of trainable parameters.

Han et al. conducted a comprehensive study, as documented in \cite{precip_models_comparison}, to assess the performance of various models in the radar precipitation nowcasting task. Among the models they examined were ConvLSTM and U-Net architectures. Their evaluation encompassed a period spanning from 2018 to 2020, as well as a specific case study focusing on a summer heavy rainfall event. The results of their investigation revealed a consistent trend: U-Net-based architectures consistently outperformed other models in terms of the critical success index (CSI) in both the long-term period and the context of the summer heavy rainfall event. Furthermore, their findings indicated that U-Net architectures achieved their best performance with shorter input sequences (60 minutes) compared to longer sequences (120 minutes). Additionally, they highlighted the significant influence of the most recent input frame on the accuracy of predictions.

Recent research has also demonstrated the potential for enhancing forecast accuracy by integrating different types of data sources. For instance, Ko et al. \cite{radar_and_ground} illustrated that combining radar-based precipitation data with ground-based weather data yields improved forecast performance. Additionally, Kaparakis et al. introduced WF-UNet \cite{wf-unet}, highlighting the benefits of incorporating wind radar data into deep learning models for more accurate precipitation forecasts. The approach by Kaparakis et al. involved a two-stream architecture, with each stream featuring a distinct U-Net. One stream takes precipitation maps as input, while the other processes wind speed maps. The outputs from both streams are combined, and a 1x1 3D convolution is applied, followed by linear activation to generate the final prediction. To evaluate the effectiveness of their approach, they compared it with four other models, including the persistence model, core UNet \cite{unet}, AsymmetricInceptionRes3DDR-UNet \cite{AsymmInceptionRes-3DDR-UNet}, and Broad-UNet \cite{broadUnet}, which did not incorporate wind data. Their findings conclusively demonstrated that WF-UNet outperformed all other models, highlighting the significant improvement in prediction performance achieved through the inclusion of wind data.

Generative adversarial models (GANs) also offer significant potential for enhancing precipitation nowcast, particularly for higher rain intensities and for generating clearer predictions. Generative Adversarial Networks (GANs) are a powerful method in generative machine learning, introduced by Goodfellow et al. \cite{goodfellowGAN}. GANs consist of two networks: a generator and a discriminator. The generator creates synthetic data resembling real data, while the discriminator distinguishes between real and fake samples. They train together in a competitive manner, with the generator improving its output to fool the discriminator, and the discriminator enhancing its ability to detect fakes. This adversarial process drives the generator to produce more realistic samples.

Ravuri et al. introduced a novel approach called the deep generative model of radar (DGMR) \cite{dgmr}, which demonstrated improved performance for extreme precipitation nowcasting compared to the conventional UNet framework. DGMR leverages a generator that utilizes convolutional gated recurrent units (GRUs) for enhanced modeling. Additionally, DGMR uses two discriminators, each focusing on distinct aspects of image quality assessment. One discriminator evaluates the temporal consistency, ensuring that predictions maintain a coherent progression over time. The other discriminator examines spatial consistency, ensuring that the generated images align with spatial patterns accurately. This approach contributes to more accurate and less blurry precipitation forecasts, particularly for higher rain intensities.

For any warning system, understanding the confidence level of predictions is crucial. Unlike Bayesian models, deep learning tools lack a solid mathematical foundation for assessing model uncertainty. Foldesi et al. \cite{uncertainty_comparison} investigated this issue by evaluating various methods for quantifying uncertainty in deep learning models, particularly when dealing with time-series data. Their findings highlighted two effective approaches: ensemble methods \cite{uncertainty_ensemble} and Dropout\cite{dropout}/DropConnect\cite{uncertainty_dropconnect} techniques. Ensemble methods involve training multiple copies of the same model architecture and combining their predictions. Lakshminarayanan et al. \cite{uncertainty_dropconnect} demonstrated that ensembles offer reliable uncertainty quantification properties, making them a valuable choice. 
On the other hand, the dropout approach generates a posterior distribution during testing by running the model for a set number of epochs ($k$) for each target precipitation map. An uncertainty map is then derived by calculating the variance across all $k$ precipitation maps. This method has been effectively employed in studies such as \cite{dropout_brain} and \cite{aatransunet}. DropConnect is comparable to Dropout, with the distinction that it sets weights to zero rather than activations. These methods provide valuable tools for assessing uncertainty in deep learning models, a critical aspect for the reliability of warning systems.

\section{Methods}\label{sec:methods}
We propose the GA-SmaAt-GNet architecture, a generative adversarial model that builds upon the SmaAt-UNet architecture \cite{SmaAtUnet}. The SmaAt-UNet architecture, an extension of the UNet model, enhances the original design by introducing Convolutional Block Attention Modules (CBAM) \cite{cbam} in the encoder to emphasize important features and depthwise-separable convolutions to reduce parameter count. These enhancements make SmaAt-UNet particularly well-suited for time series prediction tasks, facilitating precise value predictions for each pixel over time. The model's efficiency and attention-driven feature selection contribute to its improved performance. In what follows, we will first introduce our proposed generator architecture, SmaAt-GNet. Then we will explain the loss function and our proposed discriminator architecture.

\subsection{Proposed GA-SmaAt-GNet}\label{sec:ga-gnet}
\begin{figure*}[!htb]
  \centering

  \begin{subfigure}[b]{1\textwidth}
    \includegraphics[width=\textwidth]{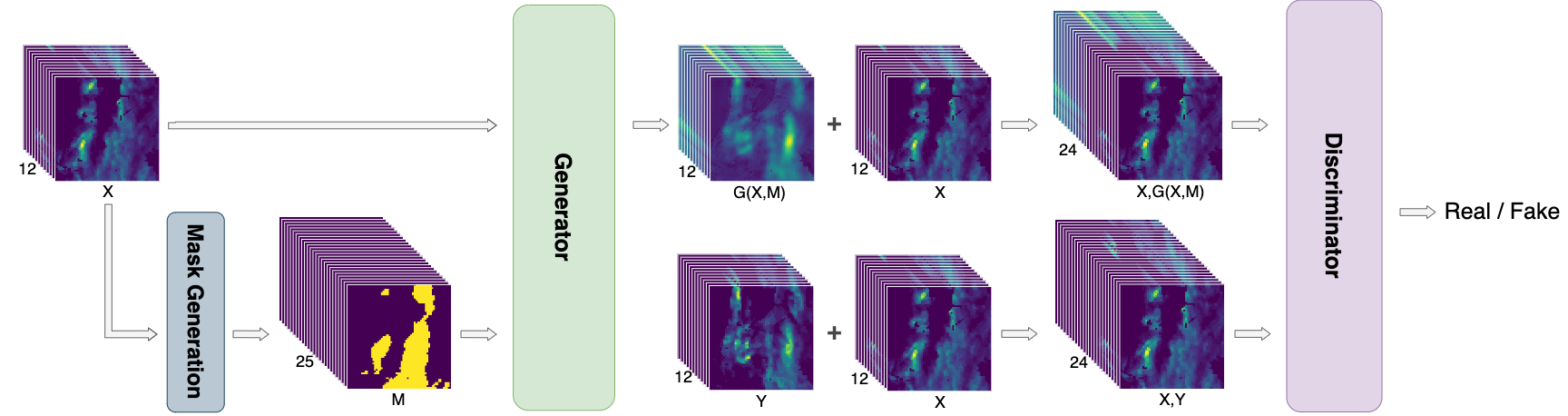}
    \caption{General overview of the GA-SmaAt-GNet architecture}
    \label{fig:gan}
  \end{subfigure}

  \bigbreak
  \bigbreak

  \begin{subfigure}[b]{0.95\textwidth}
    \includegraphics[width=\textwidth]{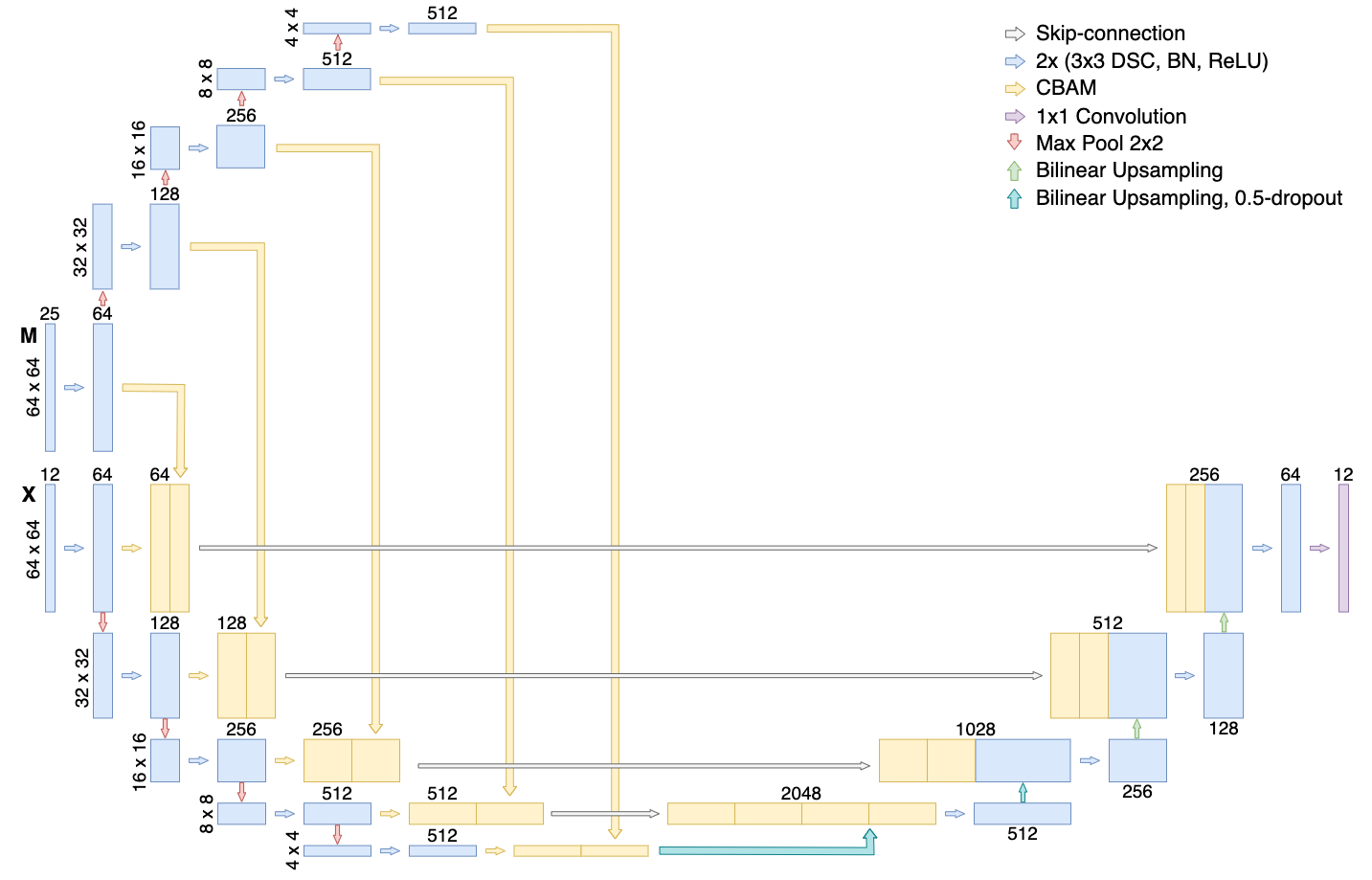}
    \caption{Generator: SmaAt-GNet}
    \label{fig:gnet}
  \end{subfigure}

  \bigbreak
  \begin{subfigure}[b]{0.95\textwidth}
    \includegraphics[width=\textwidth]{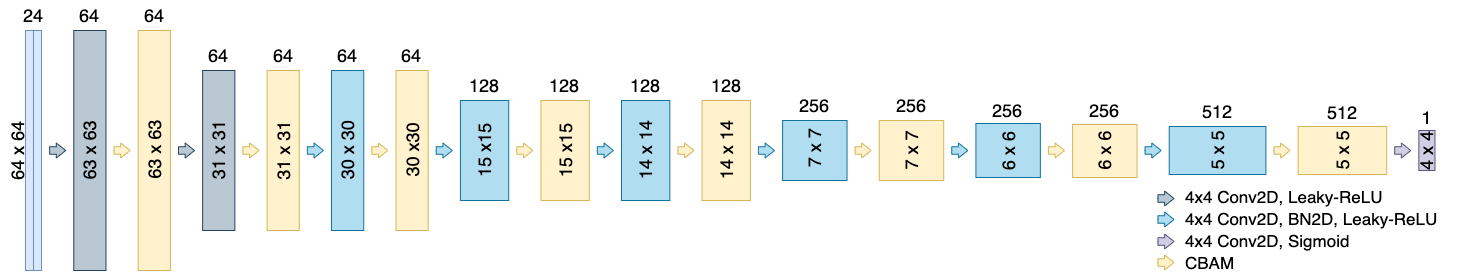}
    \caption{Attention-Augmented Discriminator}
    \label{fig:pix2pix}
  \end{subfigure}

  \caption{General overview of proposed (a) GA-SmaAt-GNet architecture with mask generation, (b) SmaAt-GNet used as generator, and (c) the Attention-Augmented Discriminator architecture.}
  \label{fig:architectures}
\end{figure*}

We first introduce SmaAt-GNet (see Fig. \ref{fig:gnet}), where ``GNet" signifies a G-shaped architecture in contrast to the U-shaped design of SmaAt-UNet. SmaAt-GNet will serve as the generator of GA-SmaAt-GNet shown in Fig. \ref{fig:architectures}. SmaAt-GNet builds upon SmaAt-UNet by incorporating a secondary encoder. This encoder facilitates the integration of an extra data source, specifically binary precipitation masks (refer to Section \ref{sec:masks} for a comprehensive explanation), into the model.
This additional encoder has the same architecture as the original SmaAt-UNet encoder, which has CBAMs as the final module at each level of the encoder.
The outputs of the CBAMs of both encoder streams are concatenated at the same level of the network.
These concatenated results are then connected to the decoder through skip connections. This design ensures that each network level can leverage features from both data sources, while also enabling attention-based feature selection within each encoder.
By sharing the same decoder, the network can uncover relationships between features from both streams, while the features themselves are learned independently. We also include two dropout layers with a dropout probability of 0.5 after the first two bilinear upsampling layers in the decoder to prevent overfitting and they also allow us to approximate the model's uncertainty as described in Section \ref{sec:epistemic}.

Our proposed GA-SmaAt-GNet (see Fig. \ref{fig:architectures}) is a conditional Generative Adversarial Network (cGAN) that merges the principles of SmaAt-GNet and the attributes of GANs to produce outputs that closely mimic real data by employing a discriminator to encourage the generator to produce real looking outputs.
A Conditional Generative Adversarial Network is a kind of GAN that incorporates conditional information during the training process. In a standard GAN, the generator learns to generate data without any specific control over the characteristics of the generated output. In contrast, a cGAN introduces conditional information, typically in the form of additional input data that guides the generation process. This additional information can be in the form of labels, class information, or in our case, precipitation maps. By providing this conditional information to the generator and discriminator during training, cGANs can be more directed in generating specific outputs.

In general, cGANs optimize performance through an adversarial loss function, defined as follows:
\begin{equation}\label{eq:l_gan}
\begin{aligned}
    \mathcal{L}_{\text{cGAN}}(G, D) =& \frac{1}{n\ell}\sum_{i=1}^{n}\sum_{j=1}^{\ell}\log [D(x_i,y_i)]_j
    +\\&\log(1 - [D(x_i,G(x_i, m_i)]_j),
\end{aligned}
\end{equation}
where $x_i$ represents the input sequence, $y_i$ is the ground truth, $m_i$ are the precipitation masks, $n$ is the number of samples, $[D(\cdot,\cdot)]_j$ indicates the $j$th element of the $4\times4$ matrix $[D(\cdot,\cdot)]$. The generator ($G$) seeks to minimize $\mathcal{L}_{\text{cGAN}}$, while the discriminator ($D$) aims to maximize it. 
Previous methodologies have demonstrated the advantage of combining the GAN objective with a regularization term, such as L1 or L2 distance \cite{pix2pix, context_enc}. In our case, we opt for L2 regularization over L1 regularization, given its greater sensitivity for extreme cases due to its squaring operation, which aligns with our focus.
\begin{equation}
\begin{aligned}
    \mathcal{L}_{\text{L2}}(G) = \frac{1}{n\kappa}\sum_{i=1}^{n}\sum_{j=1}^{\kappa}[y_i - G(x_i, m_i)]^2_j,
\end{aligned}
\end{equation}
where $y_i$ is the ground truth and  $[G(\cdot,\cdot)]_j$ indicates the $j$th element of the $12\times64\times64$ tensor $[G(\cdot,\cdot)]$.
The final objective function then becomes:
\begin{equation} \label{eg:gan_loss}
\begin{aligned}
     \min_G \max_D \mathcal{L}_{\text{GA-SmaAt-GNet}}(G, D) =& \mathcal{L}_{\text{cGAN}}(G, D) 
     + \\&\lambda  \mathcal{L}_{\text{L2}}(G),
\end{aligned}
\end{equation}
where $\lambda$ is a hyperparameter used to balance the GAN loss and regularization term. It should be noted that we apply the regularization term exclusively when optimizing the generator, while for the discriminator, the default $\mathcal{L}_{\text{cGAN}}$ from Eq. (\ref{eq:l_gan}) is optimized.

The dropout layers in the SmaAt-GNet architecture prevent the generator from becoming a simple deterministic mapping function. An alternative method to introduce randomness is by concatenating an image with random noise in the model. However, as noted in \cite{pix2pix}, a model may simply learn to ignore this noise. 

Our proposed discriminator architecture is shown in Fig. \ref{fig:pix2pix} and is an attention-augmented version of the Pix2Pix PatchGAN discriminator \cite{pix2pix}. The PatchGAN architecture consists of several convolutional layers that process the input image in a hierarchical manner. Furthermore, instead of classifying the entire image as real or fake, PatchGAN focuses on classifying small, overlapping patches of the input image. It also uses convolutions with a stride of two to gradually reduce the spatial resolution of the feature maps, helping it to capture information at different scales. 
To prevent underfitting of the discriminator caused by a lack of model complexity, we extend the PatchGAN architecture by increasing the number of convolutions. These added convolutions have a stride of one so that the spatial resolution does not decrease to quickly. Furthermore, after each convolution, a Convolutional Block Attention Module (CBAM) layer is added to focus on relevant parts of the features generated by the preceding convolutional layer.

\subsection{Training}\label{sec:training}
We conduct a comparative evaluation of our proposed GA-SmaAt-GNet and SmaAt-GNet against SmaAt-UNet \cite{SmaAtUnet}, RainNet \cite{rainnet} and the persistence model. The persistence model uses the last input image of the input sequence repeated 12 times as its prediction and RainNet is a well-established CNN that advanced precipitation nowcasting with deep learning models.

Training is performed on the precipitation dataset specified in Section \ref{sec:dataset}. The maximum epoch limit is set to 200. To enhance training efficiency and prevent overfitting, we employed the same early stopping criterion as in \cite{SmaAtUnet}, where training halts if there is no improvement on the validation set for more than 15 epochs.

For all models except RainNet \cite{rainnet}, we initialized the learning rate at 0.001, following the lines of \cite{SmaAtUnet}. For RainNet, an initial learning rate of 0.0001 was utilized, aligning with the parameters specified in the original study. To prevent overfitting of GA-SmaAt-GNet's discriminator, we adjusted the backpropagation frequency of the discriminator to once every two iterations. Additionally, a learning rate scheduler was implemented, reducing the learning rate to a tenth of its size when no improvement occurred on the validation set for four consecutive epochs. GA-SmaAt-GNet featured separate learning rate schedulers for the discriminator and generator, monitoring their respective losses on the validation set. The Adam optimizer is utilized as optimizer \cite{adam} with a batch size of 32 and Mean Squared Error (MSE) served as the loss function for SmaAt-UNet and SmaAt-GNet. For GA-SmaAt-GNet, the loss function in Eq. (\ref{eg:gan_loss}) is used with an empirically found $\lambda$-value of $10^6$. 
For RainNet, the logcosh function was employed as the loss function, consistent with the approach used in the original study \cite{rainnet}. All models are trained on an Apple M2 Ultra with 128GB of memory. The implementation of the models is available on GitHub\footnote{\url{https://github.com/EloyReulen/GA-SmaAt-GNet}} where there is also more information about pre-trained models, which are available upon request.

\subsection{Evaluation}\label{sec:evaluation}

The Mean Squared Error (MSE) is used as the standard metric for evaluating the regression performance of all models, a common practice in meteorological regression analyses \cite{SmaAtUnet, precip_models_comparison, broadUnet}. Additionally, its sensitivity to extreme values, due to the squared term, makes it particularly suitable for our focus on extreme precipitation. MSE is calculated as follows:
\begin{equation}
\begin{aligned}
    \text{MSE} &= \frac{1}{n\kappa}\sum^{n}_{i=1}\sum^{\kappa}_{j=1}[y_i-\hat{y}_i]^2_j,
\end{aligned}
\end{equation}
where $n$ represents the number of samples, $y_i$ is the ground truth and $\hat{y}_i$ is the model's prediction. Here, $y_i$ and $\hat{y}_i$ are tensors of size $12\times64\times64$ and $[y_i-\hat{y}_i]^2_j$ indicates the $j$th element of the tensor $[y_i-\hat{y}_i]^2$.

In addition to MSE, we evaluate model performance using various binary classification metrics. To this end, precipitation maps are transformed into binary representations by denormalizing and summing up 12 consecutive frames, where 12 is the number of output frames of the models. Subsequently, values exceeding specific thresholds (0.5, 10, and 20 mm/h) are set to one, while others are set to zero. This enables performance assessment for both higher and lower precipitation intensities. The thresholds of 0.5 mm/h (used to distinguish between rain and no rain) aligns with \cite{SmaAtUnet, convLSTM}. 
From these binary maps, true positives (TP) (prediction=1, target=1), false positives (FP) (prediction=1, target=0), true negatives (TN) (prediction=0, target=0), and false negatives (FN) (prediction=0, target=1) are calculated. Binary classification metrics used include F1 score, Critical Success Index (CSI), Heidke Skill Score (HSS), and Matthews Correlation Coefficient (MCC).

The F1 score is a metric commonly used in binary classification tasks to evaluate the model's performance by considering both precision and recall. Precision is the ratio of true positive predictions to the total predicted positives, while recall is the ratio of true positive predictions to the total actual positives. The F1 score is the harmonic mean of precision and recall, providing a balance between the two:
\begin{equation}
\begin{split}
    \text{Precision} &= \frac{TP}{TP + FP},\\
    \text{Recall} &= \frac{TP}{TP + FN},\\
    \text{F1 score} &= \frac{2 \times \text{Precision} \times \text{Recall}}{\text{Precision} + \text{Recall}},
\end{split}
\end{equation}
and it is particularly valuable in scenarios where there is an uneven distribution between classes, and both false positives and false negatives carry significance. 

The Critical Success Index,
\begin{equation}
    \text{CSI} = \frac{TP}{TP+FN+FP},
\end{equation}
is fundamental in evaluating the performance of detection and classification systems and is often used in meteorology. It provides a comprehensive evaluation, considering both correct and incorrect classifications.

Heidke Skill Score, 
\begin{equation}
\begin{split}
    & \text{HSS} = \\
    & \frac{TP \times TN - FP \times FN}{(TP + FN)(FN + TN) + (TP + FP)(FP + TN)},
\end{split}
\end{equation}
is also a valuable metric for evaluating the skill of binary classification models, providing a more nuanced assessment than simple accuracy. It is especially useful in scenarios where both positive and negative outcomes are important, and it helps to understand the model's performance beyond what could be achieved by chance alone, as a value smaller than 0 for this metric indicates that it performs worse than a random guessing model.

 Additionally, the Matthews Correlation Coefficient,
 \begin{equation}
 \begin{split}
    & \text{MCC} = \\
    & \frac{TP \times TN - FP \times FN}{\sqrt{(TP + FP)(TP + FN)(TN + FP)(TN + FN)}},
\end{split}
\end{equation}
is a metric that is especially useful in situations where class imbalances are present. As it offers a balanced assessment of a model's performance, considering both correct and incorrect predictions.

\section{Experiments} \label{sec:experiments}
\subsection{Dataset}\label{sec:dataset}
\begin{figure*}
    \includegraphics[width=\textwidth]{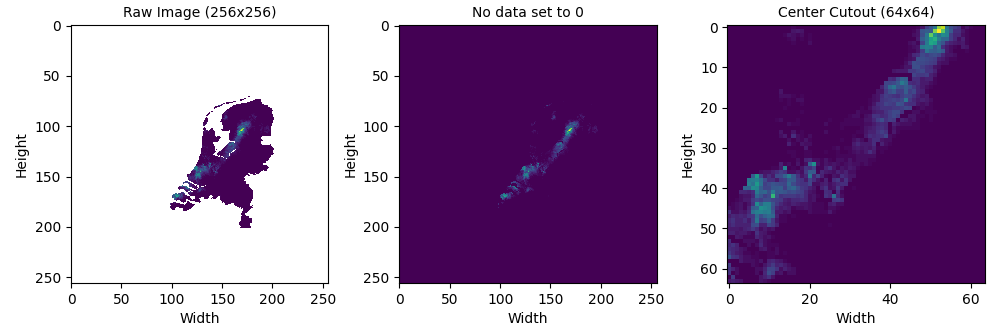}
    \caption{Preparation of the raw precipitation data. All pixels where there is no data available are set to zero and a center 64 by 64 cutout is taken from the center of the land mass of the Netherlands.}
    \label{fig:data_prep}
\end{figure*}
We train and test the models on 25 years of precipitation data from the Royal Netherlands Meteorological Institute (Koninklijk Nederlands Meteorologisch Instituut, KNMI\footnote{\url{https://www.knmi.nl/home}}). The dataset is a climatological radar rainfall dataset of 5 minute precipitation depths at a 2.4-km grid, which have been adjusted with validated and complete rain gauge data from both KNMI rain gauge networks. The first network is an automated rain gauge network of 32 stations and the second network is a manual rain gauge network of about 325 stations. The collected dataset contains data from 6 February 1998 8:05 till 31 December 2022 23:55. Of all 25 years of data, years 1998-2016 are used for the training set and years 2017-2022 are used for the test set.

Each value in the precipitation map is the precipitation dept in hundredths of millimeters (so that the values are integers). To obtain precipitation values in mm/h pixels must be multiplied by 0.01 and 12 subsequent frames must be summed up.

The raw images in the dataset have a dimension of 256 by 256 pixels that contain precipitation data for a geographical area bounded by a north latitude of 53.7, an east longitude of 7.4, a south latitude of 50.6 and a west longitude of 3.2, which includes the entire Netherlands. Only pixels that specify an area on the land mass of the Netherlands contain data. For this reason, all pixels with no data are set to 0 and a center cutout of 64 by 64 pixels is taken from the land mass of the Netherlands as shown in Fig. \ref{fig:data_prep}. The data is normalized by dividing both the values in the training and test set by the highest value found in the training set. Extremely high values can occur in the data. These are mostly non-meteorological echos. Non-meteorological echos occur when radar bundles are redirected back to the surface by large temperature and humidity gradients. To filter out most non-meteorological echos all pixels with a year sum larger than 1300 mm or 24 hour sum larger than 174 mm can be discarded.

The input for the models is a sequence of 12 precipitation maps which corresponds to one hour of precipitation data. This input length allows for a fair comparison with SmaAt-UNet \cite{SmaAtUnet} which also used 12 precipitation maps as input. The output consists of the next 12 precipitation maps. As described in \cite{SmaAtUnet} many of the precipitation maps contain little to no rain data which can cause the models to become biased towards predicting zero values. To prevent this from happening, only sequences of which the output frames contain more than 50\% rainy pixels are selected similar to the approach by \cite{SmaAtUnet}. This 50\% is calculated only on the pixels that contain precipitation data. The sample sizes of the preprocessed dataset and the original dataset are compared in Table \ref{tab:data_size}. The preprocessed dataset is available upon request, for more details see the GitHub page of this paper.

\begin{table}[h]
    \small
    \centering
    \caption{Comparison of the dataset sizes of the preprocessed dataset and the original dataset}
    \begin{tabular*}{\columnwidth}{@{\extracolsep{\fill}} l c c c}
    \hline
        Required rain pixels & Train & Test & Subset \\ \hline
        0\% (original) & 1986452 & 630864 & 100\%\\
        50\% & 51021 & 15293 & 2.53\%\\ \hline
    \end{tabular*}
    \label{tab:data_size}
\end{table}

\begin{figure}[!htb]
    \centering
    \includegraphics[width=\columnwidth]{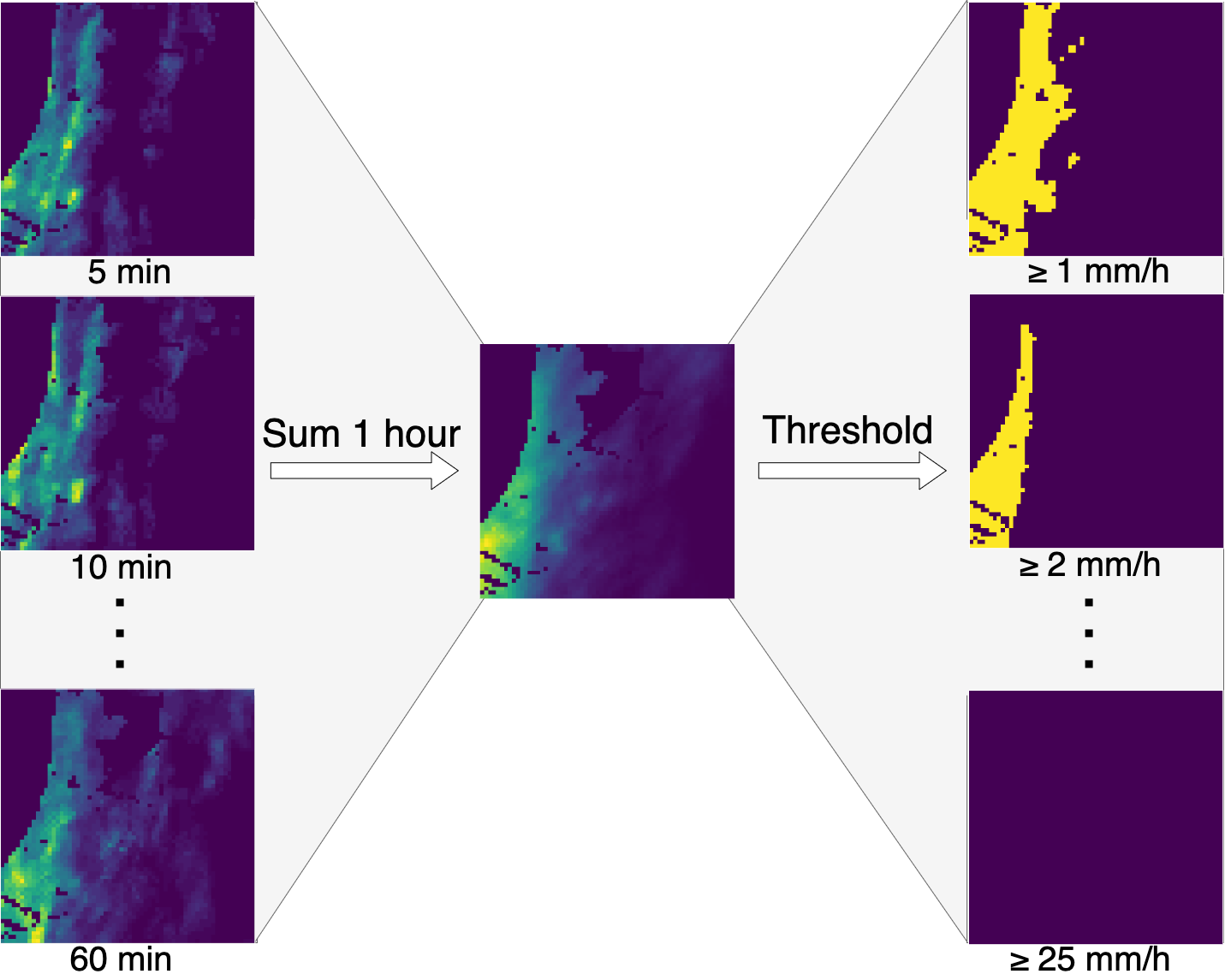}
    \caption{An example of the precipitation mask generation process. For mask generation 12 consecutive radar frames are summed up to obtain the accumulated precipitation of one hour. 25 masks are generated from this accumulated precipitation, one for each integer threshold between 1 and 25 mm/h.}
    \label{fig:mask-gen}
\end{figure}

\subsection{Precipitation masks} \label{sec:masks}
In the case of the SmaAt-GNet model, in addition to precipitation maps, a collection of binary masks is used as input. These masks are made by binarizing the accumulated rainfall over a one-hour period, as illustrated in Fig. \ref{fig:mask-gen}. Each pixel in the mask receives a value of 1 if the accumulated rainfall surpasses the specified threshold, and 0 otherwise. Each mask aligns with a distinct precipitation threshold in mm/h, ranging from 1 to 25 mm/h, resulting in a set of 25 masks for a 1 hour sequence. This augmentation supplies supplementary information to the models, pinpointing areas of specific precipitation intensities.

\section{Results and discussion} \label{sec:results}
Following the training procedure detailed in Section \ref{sec:training}, the model with the lowest validation error (MSE) for each architecture was chosen for performance evaluation on the precipitation test set. For each test case, the models are run 10 times with unfixed dropout layers. The mean of these 10 results serves as the prediction. This process makes it also possible to obtain uncertainty for the models and will be discussed in more detail in Section \ref{sec:epistemic}. Additionally, all models are compared against SmaAt-UNet \cite{SmaAtUnet}, RainNet \cite{rainnet} and the persistence model which serves as the baseline.

\subsection{Precipitation nowcasting evaluation}

\begin{figure}[!htb]
    \centering
    \includegraphics[width=\columnwidth]{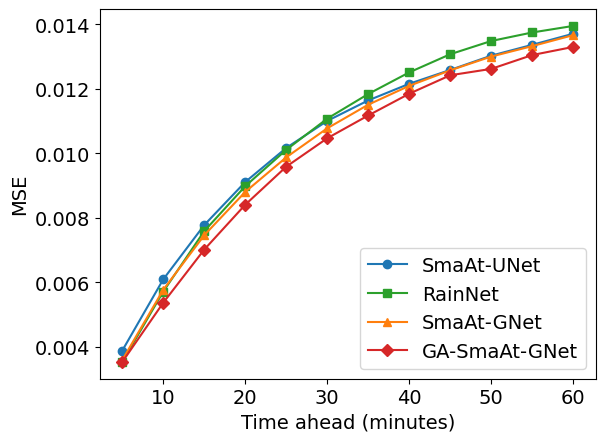}
    \caption{MSE per time step calculated on the precipitation test set. The persistence model is not shown to improve visibility.}
    \label{fig:mse_steps}
\end{figure}

The obtained results on the precipitation test set are presented in Table \ref{tab:binary} and Fig. \ref{fig:mse_steps}. It is important to note that these results are computed after denormalization, similar to those outlined by \cite{SmaAtUnet}. The results in Table \ref{tab:binary} reveal that all models perform significantly better than the persistence model in terms of Mean Squared Error (MSE). Furthermore, we observe that the introduced SmaAt-GNet architecture leads to a slight reduction in MSE on the test set when compared to SmaAt-UNet and RainNet. However, the lowest MSE of all models is observed for GA-SmaAt-GNet. 

In addition to the overall MSE on the test set, we computed MSE per predicted time step, as illustrated in Fig. \ref{fig:mse_steps}. It is evident that all models exhibit similar behavior as time progresses, i.e., MSE increases for predictions further into the future. However, when comparing SmaAt-GNet to SmaAt-UNet, it becomes clear that the overall performance gain of SmaAt-GNet can primarily be attributed to the first half-hour of the prediction. Another notable observation is that RainNet performs quite well for the initial 15 minutes of the prediction as it is only outperformed by GA-SmaAt-GNet, but as the lead time increases it is outperformed by all other models.

\begin{table*}[htb]
    \centering
    \caption{MSE, and several binary metrics calculated on the test set with different thresholds. A $\uparrow$ means that higher values for this metric are better while a $\downarrow$ means that lower values are better for this metric.}
    \begin{tabular*}{\linewidth}{@{\extracolsep{\fill}} c|c| c c c c c}
    \hline
        Threshold &  Model & MSE $\downarrow$ & F1 score $\uparrow$ &  CSI $\uparrow$ &  HSS $\uparrow$ &  MCC $\uparrow$ \\ \hline
        \multirow{4}{*}{$\geq$ 0.5 mm/h} & Persistence & 0.02297 & 0.79121 & 0.65454&0.30640&0.62084\\
        ~ & RainNet	& 0.01046 & \textbf{0.89658} & \textbf{0.81255} & \textbf{0.39190} & \textbf{0.78581} \\ 
        ~ & SmaAt-UNet & 0.01038 & 0.88785 & 0.79832 & 0.38372	& 0.76859 \\ 
        ~ & SmaAt-GNet & 0.01021& 0.89108 & 0.80356 & 0.38464 & 0.77305 \\ 
        ~ & GA-SmaAt-GNet & \textbf{0.00990} & 0.88470 & 0.79324 & 0.37375 & 0.75842 \\ \hline
        
        \multirow{4}{*}{$\geq$ 10 mm/h} & Persistence &- & 0.14543&0.07842&0.07179&0.16843 \\
        ~ & RainNet & - & 0.18330 & 0.10089 & 0.09133 & 0.20556\\
        ~ & SmaAt-UNet &- & 0.24236 & 0.13789 & 0.12093 & 0.28717 \\
        ~ & SmaAt-GNet &- & 0.26871 & 0.15521 & 0.13408 & 0.30433 \\
        ~ & GA-SmaAt-GNet &- & \textbf{0.30762} & \textbf{0.18177}  & \textbf{0.15352} & \textbf{0.33612} \\ \hline

        \multirow{4}{*}{$\geq$ 20 mm/h} & Persistence &- & 0.01944&0.00982&0.00969&0.04146\\
        ~ & RainNet & - & 0.01776 & 0.00896 & 0.00887 & 0.02166 \\
        ~ & SmaAt-UNet &- & 0.00082 & 0.00041  & 0.00041 & 0.00280 \\
        ~ & SmaAt-GNet &- & 0.05475 & 0.02815  & 0.02737 & 0.09524 \\
        ~ & GA-SmaAt-GNet &- & \textbf{0.08264} & \textbf{0.04310} & \textbf{0.04131} & \textbf{0.10266} \\ \hline
        
    \end{tabular*}
    \raggedright
    \textbf{Note:} A - indicates that the value is the same as for the previous threshold. This is the case for MSE as it is not calculated on binarized data.
    \label{tab:binary}
\end{table*}

\begin{figure*}[!htb]
    \centering
    \begin{subfigure}{0.33\textwidth}
        \includegraphics[width=\textwidth]{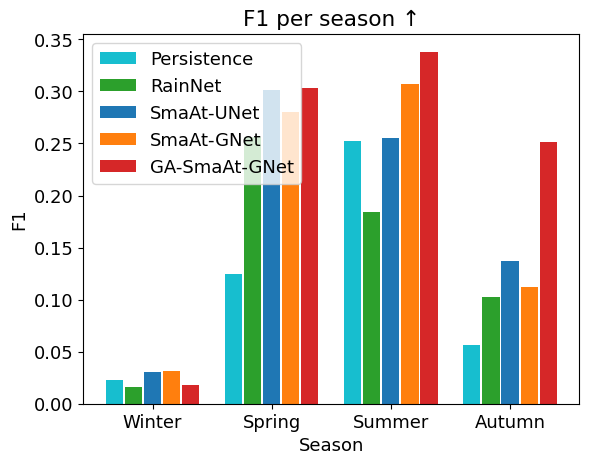}
        \caption{}
    \end{subfigure}
    \begin{subfigure}{0.33\textwidth}
        \includegraphics[width=\textwidth]{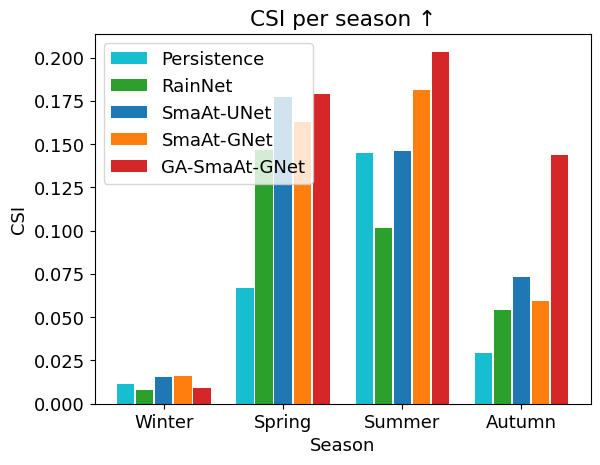}
        \caption{}
    \end{subfigure}
    \medbreak
    \begin{subfigure}{0.33\textwidth}
        \includegraphics[width=\textwidth]{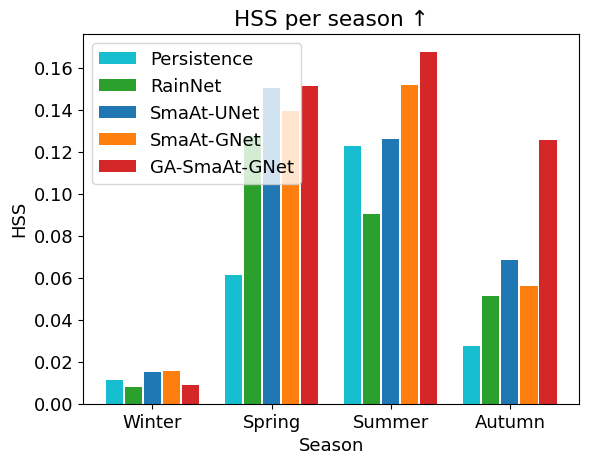}
        \caption{}
    \end{subfigure}
    \begin{subfigure}{0.33\textwidth}
        \includegraphics[width=\textwidth]{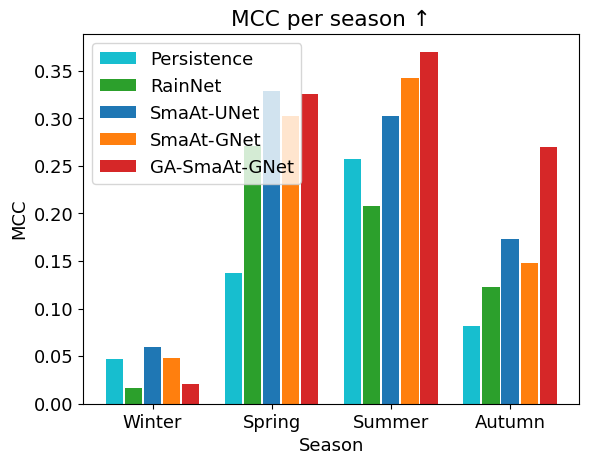}
        \caption{}
    \end{subfigure}
    \caption{Binary metrics per season for a threshold of 10 mm/h calculated on the test set. Given a specific metric, a $\uparrow$ indicates that higher values for that metric are better.}
    \label{fig:binary_season}
\end{figure*}

Furthermore, we computed various binary metrics, as detailed in Section \ref{sec:evaluation}, to assess model performance across different precipitation intensities. The results presented in Table \ref{tab:binary} reveal that different models perform better for different thresholds.

For a threshold of 0.5 mm/h, RainNet demonstrates superior performance in all binary metrics, closely followed by SmaAt-GNet, SmaAt-UNet and GA-SmaAt-GNet in that order. GA-SmaAt-GNet performs similar to the SmaAt-UNet in terms of F1 and CSI and slightly worse in terms of HSS and MCC. However, all models significantly outperform the persistence model at this threshold.

At a heavier precipitation intensity with a threshold of 10 mm/h, the persistence model performs exceptionally poor for all binary metrics. 
Among the deep learning models, RainNet performs the worst across all metrics at this threshold. Furthermore, SmaAt-GNet demonstrates an improvement over SmaAt-UNet for all metrics at this threshold. GA-SmaAt-GNet exhibits superior performance compared to all other models, outperforming them for all metrics.

In the scenario of extreme precipitation, defined by an accumulated rainfall exceeding 20 mm in an hour, one can observe that SmaAt-UNet and RainNet perform worse than the persistence model for all metrics. This may be due to high spatial smoothing, leading to underestimation of precipitation in extreme cases. This effect could arise from minimizing an objective function like Mean Squared Error (MSE) or logcosh, which tends to smooth out high-intensity values. The persistence model does not experience this issue since it simply repeats the precipitation intensity of the input. GA-SmaAt-GNet demonstrates superior performance for all binary metrics. In addition, we note that SmaAt-GNet significantly outperforms the persistence model and SmaAt-UNet in terms all binary metrics, however it is not able to reach the level of performance of the GA-SmaAt-GNet.
\begin{figure*}[!h!t!b]
  \centering
  \includegraphics[width=1\textwidth]{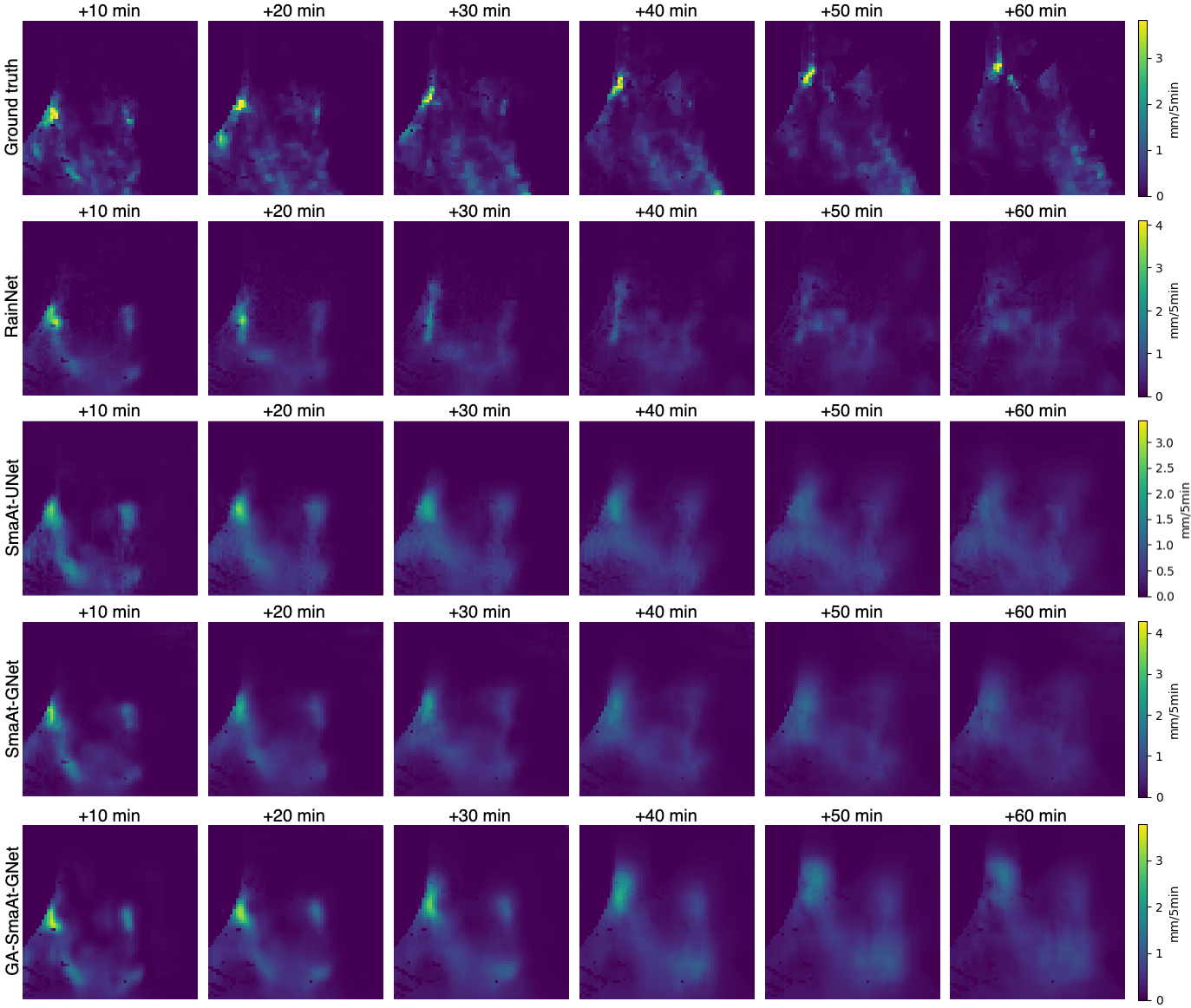}
  \caption{Example of precipitation nowcasting on a case with more than 20 mm of rain in one hour using the examined models. 
  }
  \label{fig:heavy_example}
\end{figure*}

Furthermore, we compute binary metrics per season to provide insights into model performance across different seasons. Fig. \ref{fig:binary_season} illustrates the binary metrics per season for a precipitation intensity of 10 mm/h. This threshold was chosen as no significant differences were observed for threshold of 0.5 mm/h, and for a threshold of 20 mm/h some seasons experienced a shortage of positive cases, making it less suitable for meaningful comparisons. We note that across all metrics, the tested models tend to perform less effectively in winter compared to other the other seasons. This can possibly be attributed to the scarcity of positive heavy precipitation cases in the training set during winter. Moreover, heavy precipitation in winter tends to have a different nature, being characterized by prolonged and widespread events whereas heavy summer precipitation is often local and short-lived \cite{kmni_extreme}. 
Among the deep learning models, RainNet consistently demonstrates the lowest performance across all metrics in every season. Notably, it is even outperformed by the persistence model in winter and summer. When comparing SmaAt-UNet and SmaAt-GNet, we note an interesting observation. SmaAt-UNet exhibits slightly superior performance in spring and autumn for all metrics, while SmaAt-GNet notably outperforms in summer. This is particularly interesting as summer is the season marked by the most intense rain showers \cite{kmni_extreme}. However, the standout performer in Fig. \ref{fig:binary_season} is GA-SmaAt-GNet which significantly outperforms all models for most metrics in all seasons with the exception of winter. 

\begin{figure}[!htb]
    \centering
    \includegraphics[width=\columnwidth]{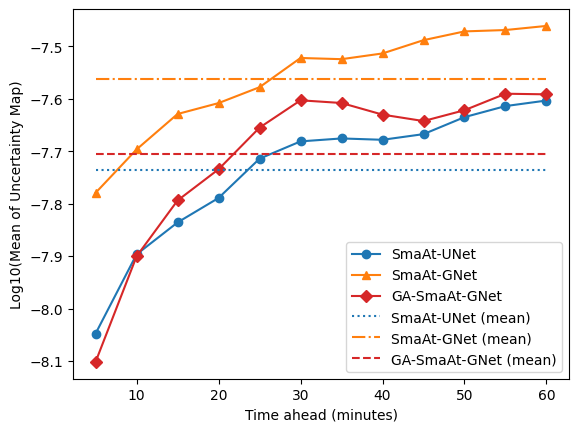}
    \caption{Epistemic uncertainty quantification results per time step calculated on the precipitation test set.
    The mean epistemic uncertainties over all time steps are indicated by the dotted lines. The persistence model is not shown as it has no uncertainty. RainNet is not included as it utilizes a different dropout technique than the other models.}
    \label{fig:uncer_mean_steps}
\end{figure}

\begin{figure*}[!ht!b]
     \begin{minipage}{.29\textwidth}
        \begin{subfigure}[h]{0.9\textwidth}
        \centering
        \includegraphics[width=\columnwidth]{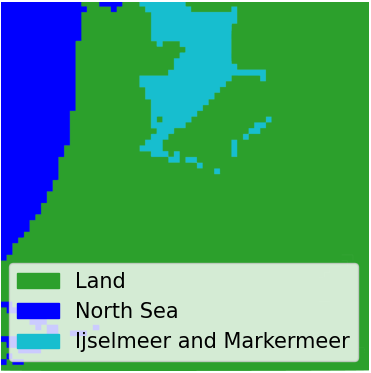}
        \caption{Locations of the North Sea, Markermeer and Ijselmeer.}
        \label{fig:map}
        \end{subfigure}
    \end{minipage}
    \begin{minipage}{.69\textwidth}
    \centering
    \begin{subfigure}[h]{0.85\textwidth}
        \centering
        \includegraphics[width=\columnwidth]{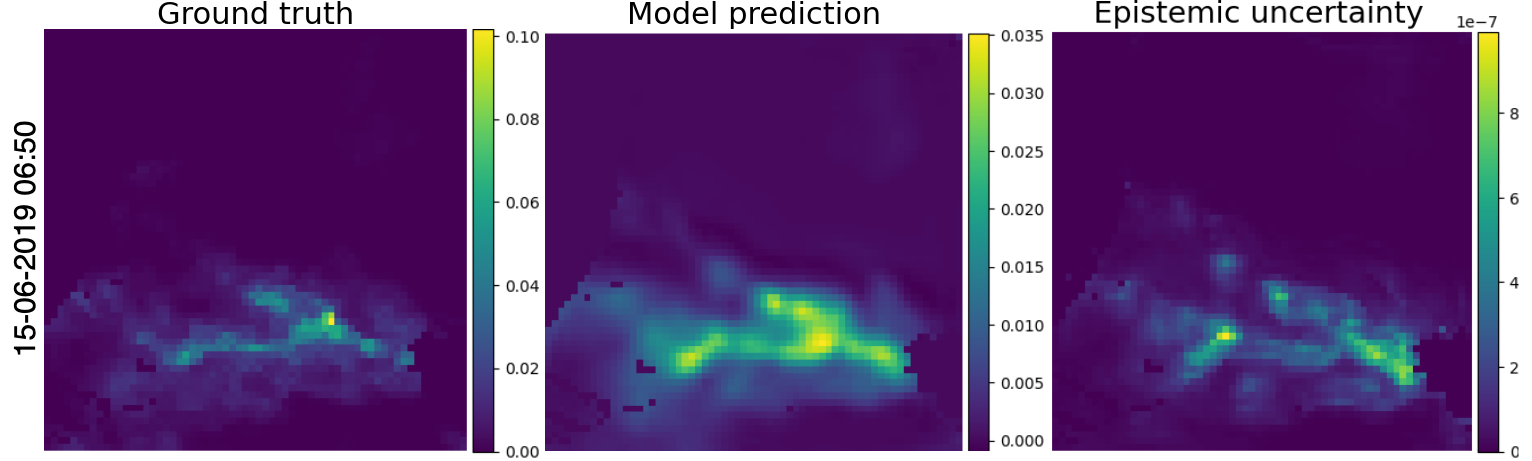}
        \caption{10 min ahead}
        \label{fig:ex_epistemic_10}
    \end{subfigure}
    \medbreak
    \begin{subfigure}[h]{0.85\textwidth}
        \centering
        \includegraphics[width=\columnwidth]{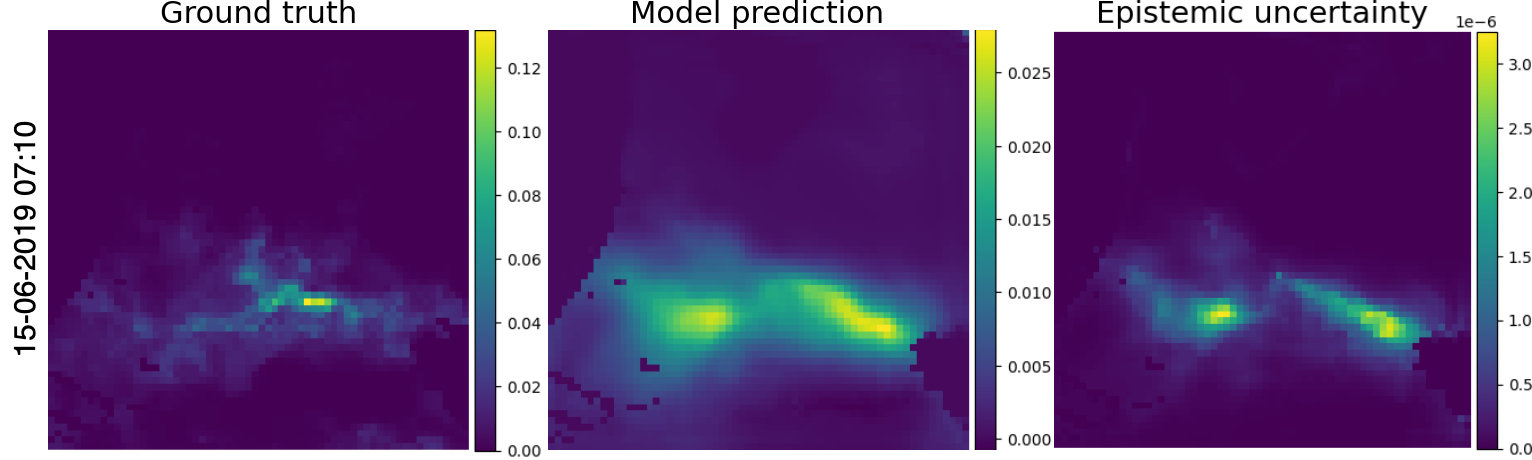}
        \caption{30 minutes ahead}
        \label{fig:ex_epistemic_30}
    \end{subfigure}
    \medbreak
    \begin{subfigure}[h]{0.85\textwidth}
        \centering
        \includegraphics[width=\columnwidth]{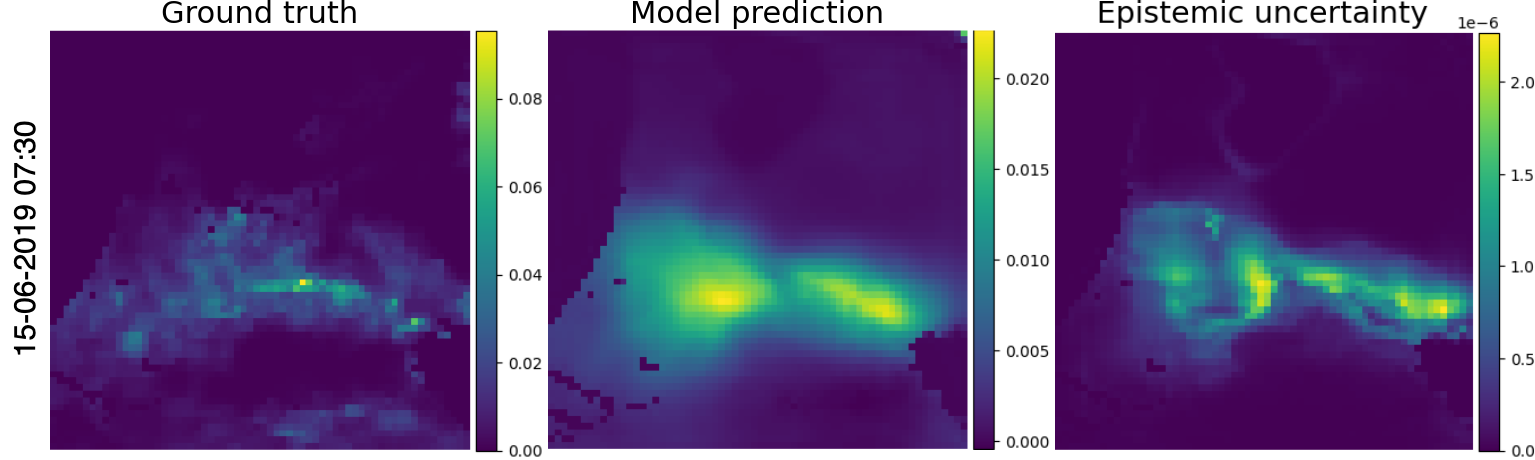}
        \caption{50 minutes ahead}
        \label{fig:ex_epistemic_50}
    \end{subfigure}
    \end{minipage}
    \caption{Epistemic uncertainty of the GA-SmaAt-GNet model for 10, 30 and 50 minutes ahead. From left to right the images are the ground truth, model prediction and epistemic uncertainty respectively. The map in (a) indicates areas of relevance for our analysis of epistemic uncertainty. Note that the values in this figure are not denormalized.}
    \label{fig:ex_epistemic}
\end{figure*}

We also provide a visual comparison of the models on a specific case from the test set where the accumulated rainfall exceeded 20 mm. The model outputs for this case are depicted in Fig. \ref{fig:heavy_example}. As anticipated, when compared with the ground truth, all models exhibit decreased accuracy as the predicted frame extends further into the future. 

When analyzing the predictions made by RainNet, we notice that they exhibit more structural detail compared to those generated by other models. This enhanced detail is likely a result of the different loss function used during RainNet's training process. Additionally, we observe that the intensity of high-precipitation area tends to decrease quite rapidly in RainNet's prediction.

Upon closer examination of the last two frames predicted by SmaAt-UNet, we note the following observation: it overestimates the rain intensity in the bottom left corner more than the other models. Additionally, in the prediction, the high intensity rain cluster spreads out over time and is almost unnoticeable in the prediction at the 60 minute mark, while in the ground truth the intensity stays the roughly same. 

Moving to SmaAt-GNet's predictions, the spread of prediction is less pronounced than that of SmaAt-UNet, but it fails to fully capture the upward movement of the high-intensity rain cluster evident in the ground truth.

The standout performer in capturing this movement is GA-SmaAt-GNet. The predictions of GA-SmaAt-GNet for 40, 50, and 60 minutes ahead showcase the high-intensity rain cluster moving significantly farther north than predictions from other models, aligning more closely with the ground truth. Additionally, for the first half hour, the prediction of the intensity of the high intensity rain cluster is sharper and the intensity resembles that of the ground truth more closely. Furthermore, the shape of the entire precipitation field is more similar to the ground truth than the other model, which is most noticeable for 50 and 60 minutes ahead.

\subsection{Uncertainty analysis}
\begin{figure*}[!htb]
    \centering
    \hspace*{-0.4 cm}
    \begin{subfigure}{0.79\textwidth}
        \includegraphics[width=1\columnwidth]{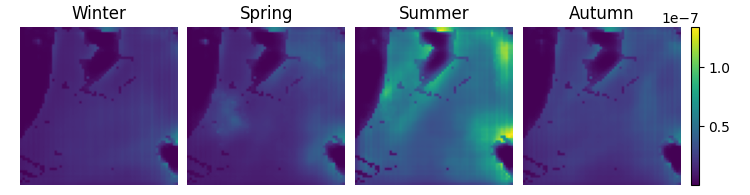}
        \caption{Mean epistemic uncertainty per season}
        \label{fig:epistemic_season}
    \end{subfigure}
    \medbreak
    \begin{subfigure}{0.8\textwidth}
        \includegraphics[width=1\columnwidth]{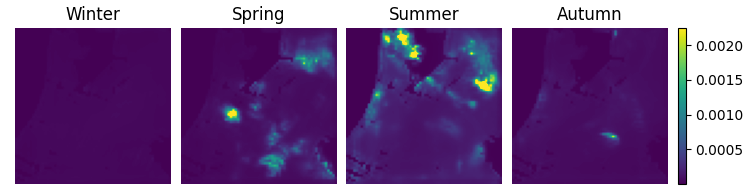}
        \caption{Mean aleatoric uncertainty per season}
        \label{fig:aleatoric_season}
    \end{subfigure}
    \medbreak
    \begin{subfigure}{0.8\textwidth}
        \includegraphics[width=1\columnwidth]{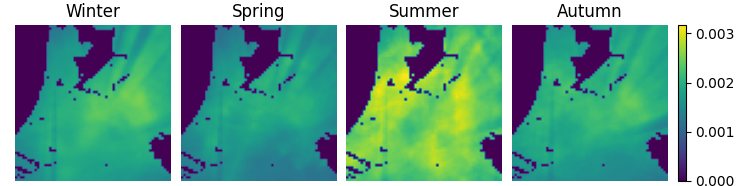}
        \caption{Mean precipitation per season}
        \label{fig:avg_precip_season}
    \end{subfigure}
    \caption{The obtained mean epistemic uncertainty, mean aleatoric uncertainty and mean precipitation per season calculated on the test set. Note that here the values are not denormalized.}
    \label{fig:al_precip_season}
\end{figure*}

In Bayesian modeling the consideration of uncertainty is critical and it is present in two primary forms \cite{uncertanties}. First, aleatoric uncertainty which is the inherent noise present in observations, such as sensor noise. This form of uncertainty persists even when the volume of data is increased. Aleatoric uncertainty is further classified into homoscedastic uncertainty, which remains constant across different inputs, and heteroscedastic uncertainty, where the degree of uncertainty depends on the inputs to the model. Heteroscedastic aleatoric uncertainty holds particular significance in computer vision applications. 

Secondly, epistemic uncertainty, often referred to as model uncertainty, addresses the uncertainty embedded in the model parameters. Unlike aleatoric uncertainty, epistemic uncertainty has the potential to decrease when expanding the dataset.

\subsubsection{Epistemic uncertainty}\label{sec:epistemic}
We approximate epistemic uncertainty by employing test-time dropout (TTD), a method used for estimating Bayesian inference, as demonstrated by \cite{aatransunet, dropout_brain}. With TTD, dropout layers are not fixed at test time like they normally would be. This intentional randomness during testing provides insights into the model's uncertainty. We perform TTD for $k=10$ epochs, generating a posterior distribution which results in 10 sets of 12 precipitation maps. Subsequently, we calculate uncertainty maps by determining the variance across the 10 precipitation maps for each of the 12 predicted time steps, resulting in 12 uncertainty maps, each corresponding to a predicted time step. If, instead of variance, we calculate the mean, we obtain the model's prediction. Additionally, if we take the mean of all uncertainty maps generated by the model we get an overall measure of uncertainty on the entire test set.

Upon examining the temporal evolution of epistemic uncertainty in Fig. \ref{fig:uncer_mean_steps}, it becomes evident that, across all models, epistemic uncertainty tends to rise as predictions project further into the future. Notably, a point of interest emerges when comparing GA-SmaAt-GNet and SmaAt-UNet: despite GA-SmaAt-GNet starting with higher certainty at 5 minutes ahead, its uncertainty escalates more rapidly than that of SmaAt-UNet. However, in the time span from 30 to 45 minutes ahead, GA-SmaAt-GNet's epistemic uncertainty experiences a slight reduction, bringing it closer to that of SmaAt-UNet.

Besides evaluating the epistemic uncertainty for all models on the test set, we also visually inspect the epistemic uncertainty for a particular case, with a specific emphasis on the proposed GA-SmaAt-GNet model. Analyzing the epistemic uncertainty in Fig. \ref{fig:ex_epistemic}, we observe that the model tends to exhibit higher uncertainty in regions where it predicts more intense rainfall. Additionally, the area of uncertainty expands as the predictions become more spread out, particularly noticeable in the transition from 30 to 50 minutes ahead. For predictions further into the future, particularly evident at 50 minutes ahead in Fig. \ref{fig:ex_epistemic_50}, the model displays slightly higher uncertainty around the edges of water. In specific areas of the top half of the image, where the uncertainty is generally low, a faint line of increased uncertainty appears along the northern coast and the edges of the Ijselmeer and Markermeer (see Fig. \ref{fig:map}). This phenomenon could be attributed to the absence of data above water in the dataset. Consequently, rain may abruptly appear and disappear at the water's edges, potentially explaining the model's uncertainty in these specific regions.

The mean epistemic uncertainties for all models on the test are indicated by the dotted lines in Fig. \ref{fig:uncer_mean_steps}. The results reveal that SmaAt-UNet is the most certain of its predictions closely followed by GA-SmaAt-GNet. SmaAt-GNet, conversely, showcases the highest uncertainty in its predictions.

\begin{figure}[!htb]
    \centering
    \begin{subfigure}[h]{\columnwidth}
        \centering
        \includegraphics[width=0.85\columnwidth]{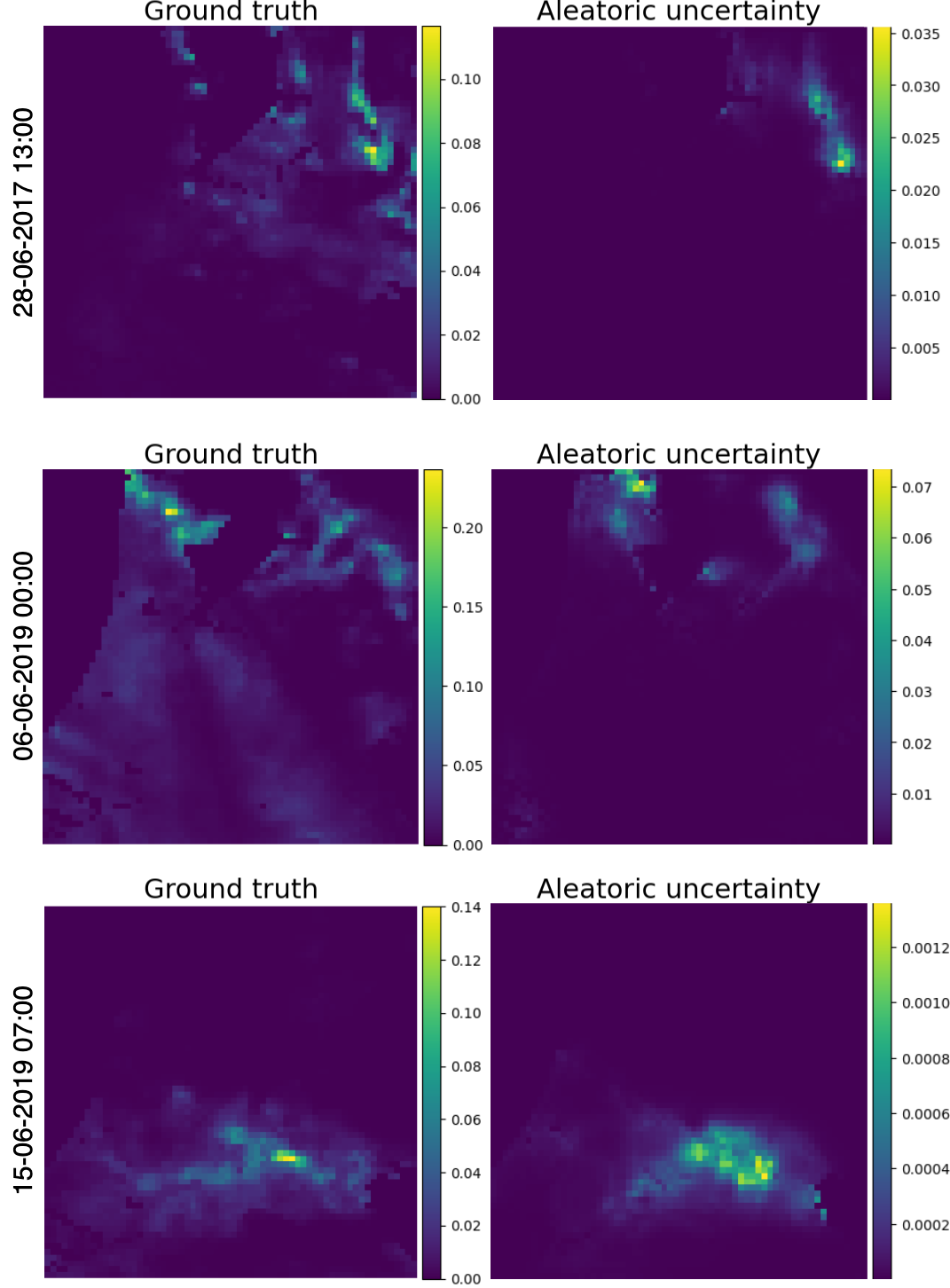}
    \end{subfigure}
    \caption{Examples of aleatoric uncertainty of the data obtained with the SmaAt-GNet model. The images on the left are the actual observed precipitation and the image on the right is the predicted aleatoric uncertainty. Note that the values in this figure are not denormalized.}
    \label{fig:ex_aleatoric}
\end{figure}

\begin{figure*}[!htb]
    \centering
    \begin{subfigure}{0.25\textwidth}
        \centering
        \includegraphics[height=10cm]{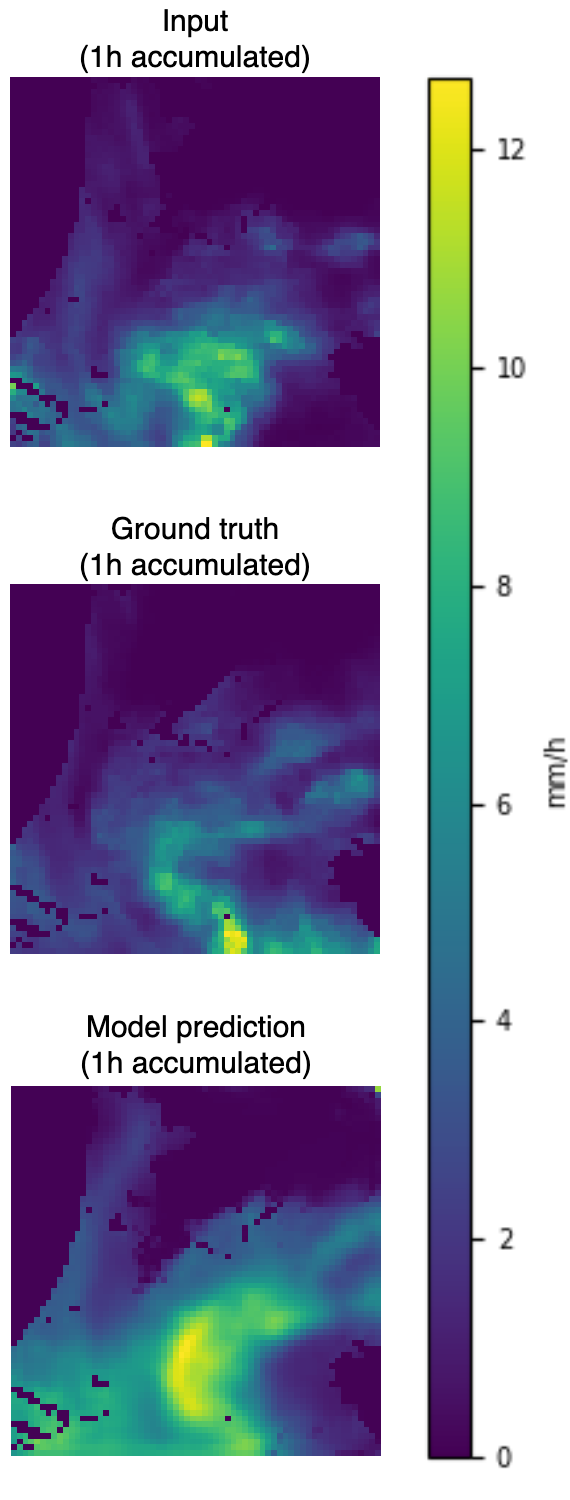}
        \caption{Input, ground truth and prediction}
        \label{fig:encoder_cam}
    \end{subfigure}
    \quad
    \begin{subfigure}{0.45\textwidth}
        \centering
        \includegraphics[height=10cm]{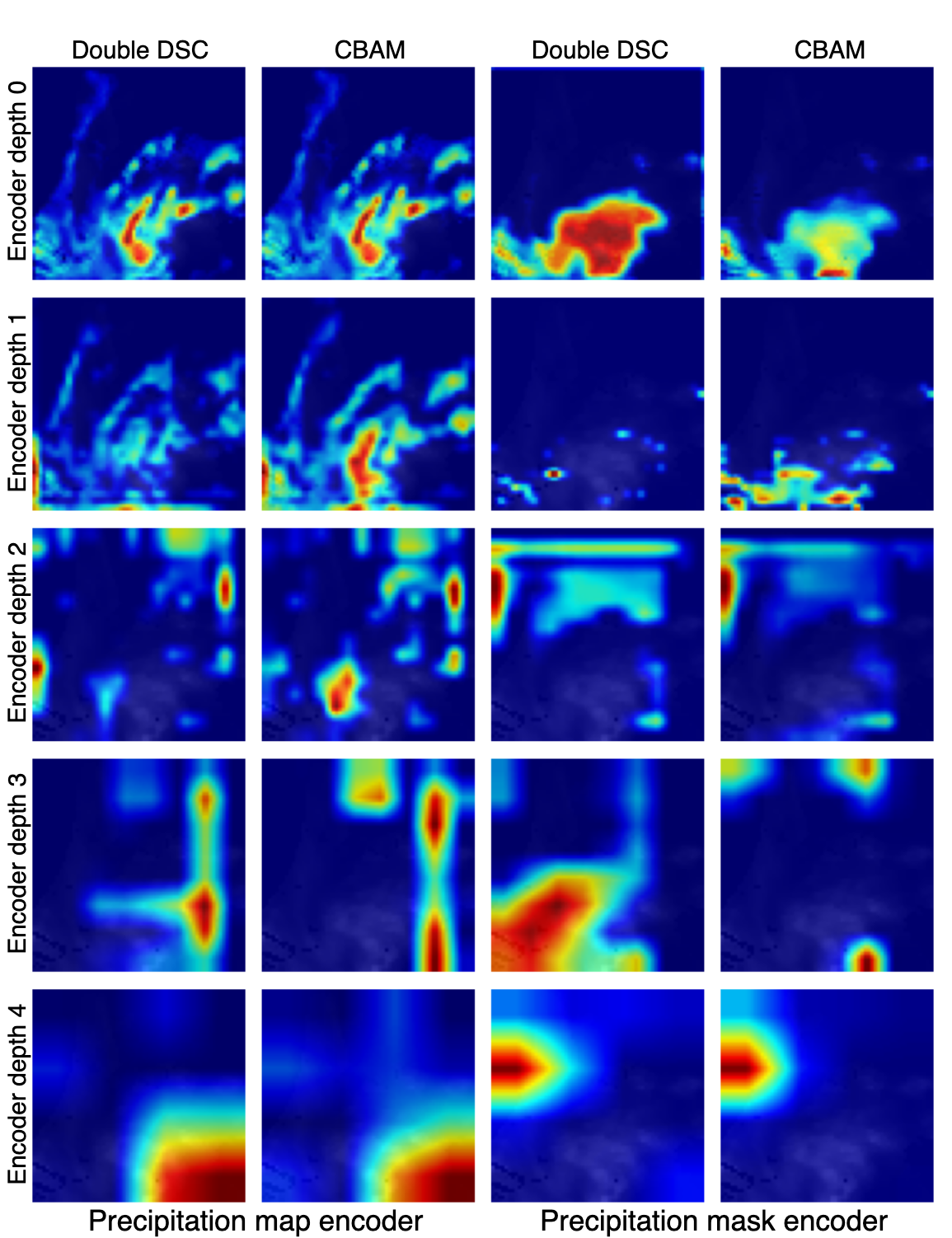}
        \caption{Encoders}
        \label{fig:encoder_cam}
    \end{subfigure}
    \quad
    \begin{subfigure}{0.2\textwidth}
        \centering
        \includegraphics[height=10cm]{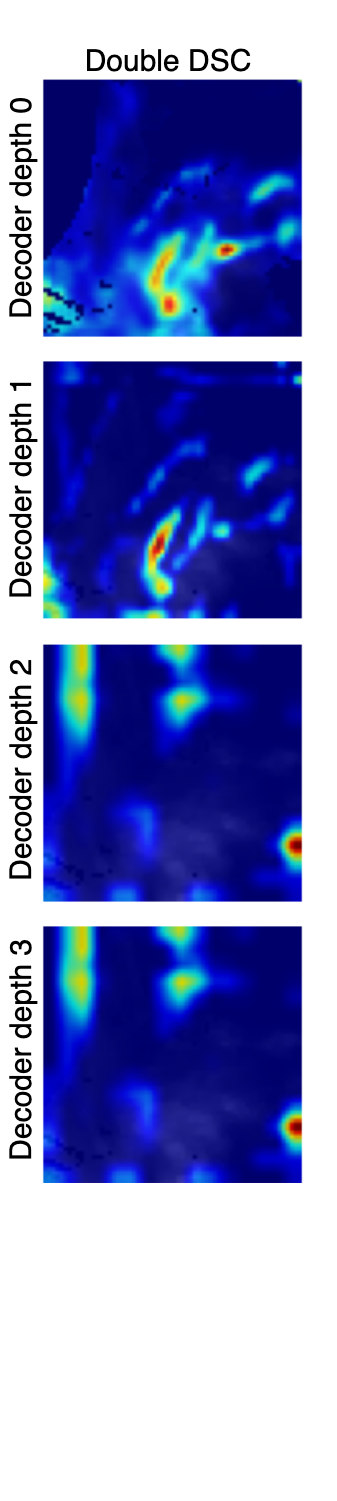}
        \caption{Decoder}
        \label{fig:decoder_cam}
    \end{subfigure}
    \caption{Activation heatmaps obtained using Grad-CAM on all pixels predicted with rain after binarization. The one hour accumulated precipitation of the input, ground truth and model prediction are shown in (a). The activation heatmaps of the encoders can be found in (b) and the activation heatmaps of the decoder can be found in (c). Each row represents  a level (depth) of the encoder or decoder of the GA-SmaAt-GNet generator. For the encoders we show activation heatmaps for the double depthwise-separable convolution (DSC) blocks and CBAM layers and for the decoder we show the activation heatmaps of the DSC blocks.}
    \label{fig:grad-cam}
\end{figure*}

Furthermore, we examine the epistemic uncertainty across seasons using the GA-SmaAt-GNet model, as illustrated in Fig. \ref{fig:al_precip_season}. An interesting finding comes to light, indicating that epistemic uncertainty reaches its highest point during the summer season. This coincides with the observation that mean precipitation is at its peak during this season, as depicted in Fig. \ref{fig:avg_precip_season}. The increased occurrence of sudden, heavy rain showers during summer, as noted in \cite{kmni_extreme}, is likely a contributing factor to the heightened epistemic uncertainty. Notably, it is intriguing to observe that, despite higher precipitation levels in winter, epistemic uncertainty tends to be lower compared to spring. This phenomenon may be attributed to the specific precipitation patterns characterizing winter, involving longer-lasting, lower-intensity rainfall events on a larger scale, as highlighted in \cite{kmni_extreme}. Consequently, these patterns suggest that epistemic uncertainty is more influenced by precipitation intensity than the overall amount, supporting our observations derived from the illustrative examples showcased in Fig. \ref{fig:ex_epistemic}.

\subsubsection{Aleatoric uncertainty}

To obtain heteroscedastic aleatoric uncertainty we require a neural network that maps the input to both an output and a measure of aleatoric uncertainty given by predicted variance, $\hat{\sigma}_i^2$, as described by Kendall et al. \cite{uncertanties}. The predicted variance $\hat{\sigma}_i^2$ is then implicitly learned from the loss function. Instead of learning $\hat{\sigma}_i^2$ directly, the log variance $s_i := \log{\hat{\sigma}_i^2}$ is learned to improve training stability by avoiding a potential division by 0. The loss function that is used to obtain aleatoric uncertainty can be defined as:
\begin{equation}
\begin{aligned}
    a_i &= \frac{1}{2} \exp(-s_i) (y_i - \hat{y}_i )^2 + \frac{1}{2} s_i, \\
    \mathcal{L}_{\text{Aleatoric,MSE}} &= 
    \frac{1}{n\kappa}\sum_{i=1}^n\sum_{j=1}^\kappa \left[a_i\right]_j,
\end{aligned}
\end{equation}
where $n$ represents the number of samples, $y_i$ is the ground truth and $\hat{y}_i$ is the prediction. Here, $a_i$, $s_i$, $y_i$ and $\hat{y}_i$ are all tensors of size $12\times64\times64$ and $[a_i]_j$ indicates the $j$th element of $[a_i]$. The predicted aleatoric uncertainty is then obtained using $\hat{\sigma}_i^2 = \exp(s_i)$.

To acquire aleatoric uncertainty, we make a slight modification to the SmaAt-GNet architecture. Specifically, we introduce a modification to include the prediction of $s_i$. We therefore add an extra 1x1 convolution at the end of the network so that the model has 2 identically shaped outputs, one for the actual predictions and one for $s_i$. This means that each pixel of the output has a corresponding value which indicates the uncertainty for that pixel.

We investigate aleatoric uncertainty through visual examples showcased in Fig. \ref{fig:ex_aleatoric}. Across all examples, a clear pattern emerges, revealing heightened uncertainty surrounding regions with higher precipitation intensity. Despite the possible reduction of epistemic uncertainty with a larger dataset, the elevated aleatoric uncertainty persists in these high-intensity regions. This increased aleatoric uncertainty poses additional challenges when predicting cases characterized by high-intensity precipitation compared to cases with lower intensity precipitation.

Additionally, we explore aleatoric uncertainty across seasons, as illustrated in Fig. \ref{fig:al_precip_season}. Similar to our observation for epistemic uncertainty, aleatoric uncertainty is also highest in summer, where mean precipitation is highest. Furthermore, in winter, mean precipitation is higher compared to spring, yet aleatoric uncertainty is lower, aligning with the observation we made for epistemic uncertainty. Thus, the patterns here also suggest that aleatoric uncertainty is more influenced by precipitation intensity than the overall amount, which aligns with our findings from the examples shown in Fig. \ref{fig:ex_aleatoric}.

\subsection{Activation heatmaps}

Explaining the outcomes of deep learning models is essential, and following the approach in \cite{sar}, we use Grad-CAM \cite{grad} to generate activation heatmaps. These heatmaps highlight the input regions influencing high values in the output, offering valuable insights into the decision-making process of the model. Originally designed for visual explanations in classification networks, Grad-CAM calculates the gradient of the class score with respect to feature map activations to determine the importance weights of neurons. It then combines these weights linearly with the forward activation maps, followed by ReLU, to produce the final outcome. To generate activation heatmaps for our nowcasting task, we convert it into an image segmentation task by binarizing the output using a threshold of 0.5 mm/h in the same way we did for our binary evaluation metrics. 

In Fig. \ref{fig:grad-cam}, we present the activation heatmaps of the double depthwise-separable convolution (DSC) blocks and Convolutional Block Attention Modules (CBAMs) at different levels (depths) of the GA-SmaAt-GNet network for a heavy precipitation case.

Examining the encoder activation heatmaps in Fig. \ref{fig:encoder_cam}, we notice distinct patterns. The precipitation map encoder's heatmaps exhibit higher activation at depths 0 and 1 in regions corresponding to elevated precipitation levels in the ground truth. In contrast, the precipitation mask encoder's heatmaps at depth 0 show increased activation in areas with higher precipitation in the model input, while at depth 1, it shows increased activation around the high precipitation area in the input. As we move to depths 2, 3, and 4, the activation areas for both encoders become more abstract and less directly linked to regions of higher precipitation, an observation also made by \cite{sar} at lower levels of their architecture.
When comparing both encoders at all depths, we see that they are generally activated in different areas, suggesting complementary contributions to the prediction. We can also observe the effects of the CBAM layers if we compare the double DSC activation maps and the CBAM activation maps. We can observe in which areas the CBAM layers emphasize and de-emphasize the double DSC activations. As an example, consider the precipitation map encoder at depth 1; after the application of the CBAM layer, the region corresponding to higher precipitation in the model's prediction exhibits increased activation. Additionally, the influence of CBAM layers is more pronounced in the precipitation mask encoder than in the precipitation map encoder, evident from the greater differences between the activation heatmaps of the DSC blocks and the CBAM layers in the precipitation mask encoder.

Examining the activation heatmaps of the DSC blocks in the decoder (Fig. \ref{fig:decoder_cam}), one can observe that at depths 3 and 2, the activation areas do not align with regions of high precipitation in the model's prediction. However, at depth 1 and 0, areas with more precipitation in the model's prediction generally exhibit heightened activation. Notably, at depth 0, the transitions in the activations of the double DSC block become more gradual compared to the activations at depth 1, resembling the final prediction closely.  This aligns logically with its position as one of the concluding layers of the network.

\section{Conclusion and future work}\label{sec:conclusion}
In this paper we introduced GA-SmaAt-GNet, a novel generative adversarial model for extreme precipitation nowcasting. GA-SmaAt-GNet uses two methodologies to more accurately nowcast extreme precipitation: the integration of precipitation masks in a novel SmaAt-GNet architecture and a generative adversarial approach with an attention-augmented discriminator. The validity and applicability of the proposed models are shown on real-life precipitation dataset of Netherlands spanning 25 years. Furthermore, we demonstrated that both aleatoric and epistemic uncertainty increase with higher precipitation intensities, emphasizing the inherent challenges associated with the task of extreme precipitation nowcasting. Finally, we utilized Grad-CAM to generate activation heatmaps for various parts of our proposed GA-SmaAt-GNet, providing valuable insights into the model's predictions. While the methodologies presented in this paper showcase promising results for extreme precipitation nowcasting, further research is essential to refine the accuracy and certainty of predictions, particularly for longer lead times. For the future work one may consider adapting other possible loss functions such as weighted loss or incorporating other weather variables and exploring the inclusion of different data sources.

\bibliographystyle{elsarticle-num}
\bibliography{references}

\begin{thebibliography}{10}
\expandafter\ifx\csname url\endcsname\relax
  \def\url#1{\texttt{#1}}\fi
\expandafter\ifx\csname urlprefix\endcsname\relax\def\urlprefix{URL }\fi
\expandafter\ifx\csname href\endcsname\relax
  \def\href#1#2{#2} \def\path#1{#1}\fi

\bibitem{nowcasting}
A.~Hering, C.~Morel, G.~Galli, S.~S{\'e}n{\'e}si, P.~Ambrosetti, M.~Boscacci, Nowcasting thunderstorms in the alpine region using a radar based adaptive thresholding scheme, in: Proceedings of ERAD, Vol.~1, 2004.

\bibitem{precip_trends_nl}
E.~Daniels, G.~Lenderink, R.~Hutjes, A.~Holtslag, Spatial precipitation patterns and trends in the netherlands during 1951--2009, International journal of climatology 34~(6) (2014) 1773--1784.

\bibitem{kmni_extreme}
J.~Beersma, H.~Hakvoort, R.~Jilderda, A.~Overeem, R.~Versteeg, Neerslagstatistiek en -reeksen voor het waterbeheer 2019, \url{https://www.stowa.nl/sites/default/files/assets/PUBLICATIES/Publicaties%202019/STOWA%202019-19%20neerslagstatistieken.pdf}.

\bibitem{extreme_precip_nl}
J.~M. Eden, S.~F. Kew, O.~Bellprat, G.~Lenderink, I.~Manola, H.~Omrani, G.~J. van Oldenborgh, Extreme precipitation in the netherlands: An event attribution case study, Weather and climate extremes 21 (2018) 90--101.

\bibitem{nwp}
P.~Bauer, A.~Thorpe, G.~Brunet, The quiet revolution of numerical weather prediction, Nature 525~(7567) (2015) 47--55.

\bibitem{deep_shared}
S.~Mehrkanoon, Deep shared representation learning for weather elements forecasting, Knowledge-Based Systems 179 (2019) 120--128.

\bibitem{alex}
A.~Krizhevsky, I.~Sutskever, G.~E. Hinton, Imagenet classification with deep convolutional neural networks, Advances in neural information processing systems 25 (2012).

\bibitem{resnet}
K.~He, X.~Zhang, S.~Ren, J.~Sun, Deep residual learning for image recognition, in: Proceedings of the IEEE conference on computer vision and pattern recognition, 2016, pp. 770--778.

\bibitem{vgg}
K.~Simonyan, A.~Zisserman, Very deep convolutional networks for large-scale image recognition, arXiv preprint arXiv:1409.1556 (2014).

\bibitem{kim2014convolutional}
Y.~Kim, Convolutional neural networks for sentence classification, arXiv preprint arXiv:1408.5882 (2014).

\bibitem{zhang2016characterlevel}
X.~Zhang, J.~Zhao, Y.~LeCun, Character-level convolutional networks for text classification, Advances in neural information processing systems 28 (2015).

\bibitem{dos}
X.~Song, N.~Wu, S.~Song, V.~Stojanovic, Switching-like event-triggered state estimation for reaction--diffusion neural networks against dos attacks, Neural Processing Letters 55~(7) (2023) 8997--9018.

\bibitem{fuzzycyberattack}
Z.~Zhang, X.~Song, X.~Sun, V.~Stojanovic, Hybrid-driven-based fuzzy secure filtering for nonlinear parabolic partial differential equation systems with cyber attacks, International Journal of Adaptive Control and Signal Processing 37~(2) (2023) 380--398.

\bibitem{convLSTM}
X.~Shi, Z.~Chen, H.~Wang, D.-Y. Yeung, W.-K. Wong, W.-c. Woo, Convolutional lstm network: A machine learning approach for precipitation nowcasting, Advances in neural information processing systems 28 (2015).

\bibitem{trajGRU}
X.~Shi, Z.~Gao, L.~Lausen, H.~Wang, D.-Y. Yeung, W.-k. Wong, W.-c. Woo, Deep learning for precipitation nowcasting: A benchmark and a new model, Advances in neural information processing systems 30 (2017).

\bibitem{stunner}
W.~Fang, L.~Pang, V.~S. Sheng, Q.~Wang, Stunner: Radar echo extrapolation model based on spatio-temporal fusion neural network, IEEE Transactions on Geoscience and Remote Sensing (2023).

\bibitem{scent}
W.~Fang, F.~Zhang, V.~S. Sheng, Y.~Ding, Scent: A new precipitation nowcasting method based on sparse correspondence and deep neural network, Neurocomputing 448 (2021) 10--20.

\bibitem{trebing2020wind}
K.~Trebing, S.~Mehrkanoon, Wind speed prediction using multidimensional convolutional neural networks, in: IEEE symposium series on computational intelligence (IEEE-SSCI), IEEE, 2020, pp. 713--720.

\bibitem{dgmr}
S.~Ravuri, K.~Lenc, M.~Willson, D.~Kangin, R.~Lam, P.~Mirowski, M.~Fitzsimons, M.~Athanassiadou, S.~Kashem, S.~Madge, et~al., Skilful precipitation nowcasting using deep generative models of radar, Nature 597~(7878) (2021) 672--677.

\bibitem{extreme_hard}
G.~Franch, D.~Nerini, M.~Pendesini, L.~Coviello, G.~Jurman, C.~Furlanello, Precipitation nowcasting with orographic enhanced stacked generalization: Improving deep learning predictions on extreme events, Atmosphere 11~(3) (2020) 267.

\bibitem{gan_mask}
R.~Wang, L.~Su, W.~K. Wong, A.~K. Lau, J.~C. Fung, Skillful radar-based heavy rainfall nowcasting using task-segmented generative adversarial network, IEEE Transactions on Geoscience and Remote Sensing (2023).

\bibitem{nowcastnet}
Y.~Zhang, M.~Long, K.~Chen, L.~Xing, R.~Jin, M.~I. Jordan, J.~Wang, Skilful nowcasting of extreme precipitation with nowcastnet, Nature 619~(7970) (2023) 526--532.

\bibitem{knmi_zwaar}
KNMI, Zware regen, \url{https://www.knmi.nl/kennis-en-datacentrum/waarschuwingen/zware-regen}.

\bibitem{SmaAtUnet}
K.~Trebing, T.~Sta{\'n}czyk, S.~Mehrkanoon, \uppercase{S}ma\uppercase{A}t-\uppercase{UN}et: Precipitation nowcasting using a small attention-unet architecture, Pattern Recognition Letters 145 (2021) 178--186.

\bibitem{cbam}
S.~Woo, J.~Park, J.-Y. Lee, I.~S. Kweon, Cbam: Convolutional block attention module, in: Proceedings of the European conference on computer vision (ECCV), 2018, pp. 3--19.

\bibitem{rainnet}
G.~Ayzel, T.~Scheffer, M.~Heistermann, Rainnet v1. 0: a convolutional neural network for radar-based precipitation nowcasting, Geoscientific Model Development 13~(6) (2020) 2631--2644.

\bibitem{unet}
O.~Ronneberger, P.~Fischer, T.~Brox, U-net: Convolutional networks for biomedical image segmentation, in: Medical Image Computing and Computer-Assisted Intervention--MICCAI 2015: 18th International Conference, Munich, Germany, October 5-9, 2015, Proceedings, Part III 18, Springer, 2015, pp. 234--241.

\bibitem{aatransunet}
Y.~Yang, S.~Mehrkanoon, \uppercase{AA}-\uppercase{T}rans\uppercase{un}et: Attention augmented transunet for nowcasting tasks, in: 2022 International Joint Conference on Neural Networks (IJCNN), IEEE, 2022, pp. 01--08.

\bibitem{sar}
M.~Renault, S.~Mehrkanoon, \uppercase{SAR-UNet}: Small attention residual unet for explainable nowcasting tasks, in: 2023 International Joint Conference on Neural Networks (IJCNN), 2023, pp. 1--8.
\newblock \href {https://doi.org/10.1109/IJCNN54540.2023.10191095} {\path{doi:10.1109/IJCNN54540.2023.10191095}}.

\bibitem{precip_models_comparison}
D.~Han, J.~Im, Y.~Shin, J.~Lee, Key factors for quantitative precipitation nowcasting using ground weather radar data based on deep learning, Geoscientific Model Development 16~(20) (2023) 5895--5914.

\bibitem{radar_and_ground}
J.~Ko, K.~Lee, H.~Hwang, K.~Shin, Deep-learning-based precipitation nowcasting with ground weather station data and radar data, in: 2022 IEEE International Conference on Data Mining Workshops (ICDMW), IEEE, 2022, pp. 1063--1070.

\bibitem{wf-unet}
C.~Kaparakis, S.~Mehrkanoon, \uppercase{WF-UN}et: Weather data fusion using 3d-unet for precipitation nowcasting, Procedia Computer Science 222 (2023) 223--232.

\bibitem{AsymmInceptionRes-3DDR-UNet}
J.~G. Fern{\'a}ndez, I.~A. Abdellaoui, S.~Mehrkanoon, Deep coastal sea elements forecasting using unet-based models, Knowledge-Based Systems 252 (2022) 109445.

\bibitem{broadUnet}
J.~G. Fern{\'a}ndez, S.~Mehrkanoon, \uppercase{B}road-\uppercase{UN}et: Multi-scale feature learning for nowcasting tasks, Neural Networks 144 (2021) 419--427.

\bibitem{goodfellowGAN}
I.~Goodfellow, J.~Pouget-Abadie, M.~Mirza, B.~Xu, D.~Warde-Farley, S.~Ozair, A.~Courville, Y.~Bengio, Generative adversarial nets, Advances in neural information processing systems 27 (2014).

\bibitem{uncertainty_comparison}
L.~Foldesi, M.~Valdenegro-Toro, Comparison of uncertainty quantification with deep learning in time series regression, arXiv preprint arXiv:2211.06233 (2022).

\bibitem{uncertainty_ensemble}
B.~Lakshminarayanan, A.~Pritzel, C.~Blundell, Simple and scalable predictive uncertainty estimation using deep ensembles, Advances in neural information processing systems 30 (2017).

\bibitem{dropout}
Y.~Gal, Z.~Ghahramani, Dropout as a bayesian approximation: Representing model uncertainty in deep learning, in: international conference on machine learning, PMLR, 2016, pp. 1050--1059.

\bibitem{uncertainty_dropconnect}
A.~Mobiny, P.~Yuan, S.~K. Moulik, N.~Garg, C.~C. Wu, H.~Van~Nguyen, Dropconnect is effective in modeling uncertainty of bayesian deep networks, Scientific reports 11~(1) (2021) 5458.

\bibitem{dropout_brain}
P.~Natekar, A.~Kori, G.~Krishnamurthi, Demystifying brain tumor segmentation networks: interpretability and uncertainty analysis, Frontiers in computational neuroscience 14 (2020) 6.

\bibitem{pix2pix}
P.~Isola, J.-Y. Zhu, T.~Zhou, A.~A. Efros, Image-to-image translation with conditional adversarial networks, in: Proceedings of the IEEE conference on computer vision and pattern recognition, 2017, pp. 1125--1134.

\bibitem{context_enc}
D.~Pathak, P.~Krahenbuhl, J.~Donahue, T.~Darrell, A.~A. Efros, Context encoders: Feature learning by inpainting, in: Proceedings of the IEEE conference on computer vision and pattern recognition, 2016, pp. 2536--2544.

\bibitem{adam}
D.~P. Kingma, J.~Ba, Adam: A method for stochastic optimization, arXiv preprint arXiv:1412.6980 (2014).

\bibitem{uncertanties}
A.~Kendall, Y.~Gal, What uncertainties do we need in bayesian deep learning for computer vision?, Advances in neural information processing systems 30 (2017).

\bibitem{grad}
R.~R. Selvaraju, M.~Cogswell, A.~Das, R.~Vedantam, D.~Parikh, D.~Batra, Grad-cam: Visual explanations from deep networks via gradient-based localization, in: Proceedings of the IEEE international conference on computer vision, 2017, pp. 618--626.

\end{thebibliography}

\end{document}